\documentclass[lettersize,journal]{IEEEtran}
\usepackage{amsmath,amsfonts}
\usepackage{algorithmic}
\usepackage{algorithm}
\usepackage{array}
\usepackage[caption=false,font=normalsize,labelfont=sf,textfont=sf]{subfig}
\usepackage{textcomp}
\usepackage{stfloats}
\usepackage{url}
\usepackage{verbatim}
\usepackage{graphicx}
\usepackage{cite}
\usepackage{times}
\usepackage{epsfig}
\usepackage{graphicx}
\usepackage{amssymb}
\usepackage{mathabx}
\usepackage{fontawesome}
\begin{document}

\title{Exploring Richer and More Accurate Information via Frequency Selection for Image Restoration}

\author{Hu Gao, Depeng Dang$^\dag$
\thanks{Hu Gao and Depeng Dang are with the School of
Artificial Intelligence, Beijing Normal University,
Beijing 100000, China (e-mail: gao\_h@mail.bnu.edu.cn, ddepeng@bnu.edu.cn).}}



\maketitle

\begin{abstract}
Image restoration aims to recover high-quality images from their corrupted counterparts.
Many existing methods primarily focus on the spatial domain, neglecting the understanding of frequency variations and ignoring the impact of implicit noise in skip connections.
In this paper, we introduce a multi-scale frequency selection network (MSFSNet) that seamlessly integrates spatial and frequency domain knowledge, selectively recovering richer and more accurate information.
Specifically, we initially capture spatial features and input them into  dynamic filter selection modules (DFS)  at different scales to integrate frequency knowledge. DFS utilizes learnable filters to generate high and low-frequency information and employs a frequency cross-attention mechanism (FCAM) to determine the most information to recover.
To learn a multi-scale and accurate set of hybrid features, we develop a skip feature fusion block (SFF) that leverages contextual features to discriminatively determine which information should be propagated in skip-connections. 
It is worth noting that our DFS and SFF are generic plug-in modules that can be directly employed in existing networks without any adjustments, leading to performance improvements.
Extensive experiments across various image restoration tasks demonstrate that our MSFSNet achieves performance that is either superior or comparable to state-of-the-art algorithms. The code and the pre-trained models are released at \url{https://github.com/Tombs98/MSFSNet_}

\end{abstract}


\section{Introduction}
\IEEEPARstart{I}{mage} restoration is a longstanding low-level vision problem that seeks to recover a latent sharp image from its corrupted counterpart. It is an commonly ill-posed inverse problem, many
conventional approaches~\cite{yang2020single, karaali2017edge, 2011Image, 2011Single} address this by explicitly incorporating various priors or hand-crafted features to constrain the solution space to natural images. Nonetheless, designing such priors proves challenging and lacks generalizability.

Recently, deep neural networks have undergone rapid development in the field of image restoration. Leveraging large-scale data, deep models like Convolutional Neural Networks (CNNs) and Transformers can implicitly learn more general priors by capturing image features, resulting in superior performance compared to conventional methods. The performance enhancement of deep learning methods is primarily attributed to their model design. Over the years, numerous network modules and functional units for image restoration have been developed by either inventing new ones or borrowing advanced modules from other domains. This has led to the creation of various models, including encoder-decoder architectures~\cite{IRNeXt,chen2022simple,2021Rethinking}, multi-stage networks~\cite{Zamir2021MPRNet, Chen_2021_CVPR, PREnet, RESCAN}, dual networks~\cite{2018LearningD, 2022Learning, 2020Refining, chen2020decomposition}, recursive residual structures~\cite{zhang2018image, 2019Real,ms9919385}, generative models~\cite{Degan,DBGAN,deganv2,mg10130403,mg9671019}, and various transformer enhancements~\cite{u2former,IDT,Zamir2021Restormer,Tsai2022Stripformer, Wang_2022_CVPR,mt10387581}.

However, the aforementioned methods concentrate on the spatial domain, without adequately considering the knowledge of frequency variation between sharp and degraded image pairs. To this end, few methods~\cite{kong2023efficient,fLi2023ICLR,f8803391,fxint2023freqsel,WACAFRN,huang2022WINNet,mf9786841,mf9917526} utilize transformation tools to explore the frequency domain.
Nevertheless, these methods primarily concentrate on capturing global information or reducing the space and time complexity of the self-attention mechanism. They fail to consider how to effectively utilize both low-frequency and high-frequency information to selectively choose the most informative frequency for recovery. 
While \cite{FSNet, SFNet, 10196308} emphasize or attenuate frequency components to select the most informative ones for recovery, they only fuse low- and high-frequency information in the channel dimension. This often results in difficulties distinguishing background details in local regions and an inability to balance overlapping scenes.

Given the information presented, a natural question arises: Is it feasible to design a network that seamlessly integrates spatial and frequency domain knowledge and adaptively selects the most informative frequency for recovery? To achieve this objective, we propose a multi-scale frequency selection network, called MSFSNet, with several key components.
Specifically, we design the multi-scale frequency selection block (MSFS) that initially capture spatial domain features by NAFBlock \cite{chen2022simple}. These spatial features are then input into two parallel dynamic filtering selection modules (DFS) with different scales to integrate frequency variation knowledge. The DFS involves two main processes: (1) Dynamically generating high and low-frequency maps using a learnable filter. (2) Utilizing the frequency cross-attention mechanism (FCAM) to discriminate which low-frequency and high-frequency information should be retained. Ultimately, the most informative component is selected for the recovery of the potentially clear image.

Additionally, existing methods where the features of the encoder and decoder are skip-connected through simple connections or additions. However, the features from the encoder may include image degradation factors, and the simple addition or concatenation process between the encoder and decoder is prone to implicit noise, thereby affecting the model's image restoration ability. To this end, we introduces a skip feature fusion block (SFF), which employs a attention mechanism to discriminatively determine the information that should propagate in skip connections. Importantly, SFF combines features from different scales while preserving their unique complementary properties, enabling our model to acquire a rich and accurate set of features.

\begin{figure*}
    \centering
    \includegraphics[width=1\linewidth]{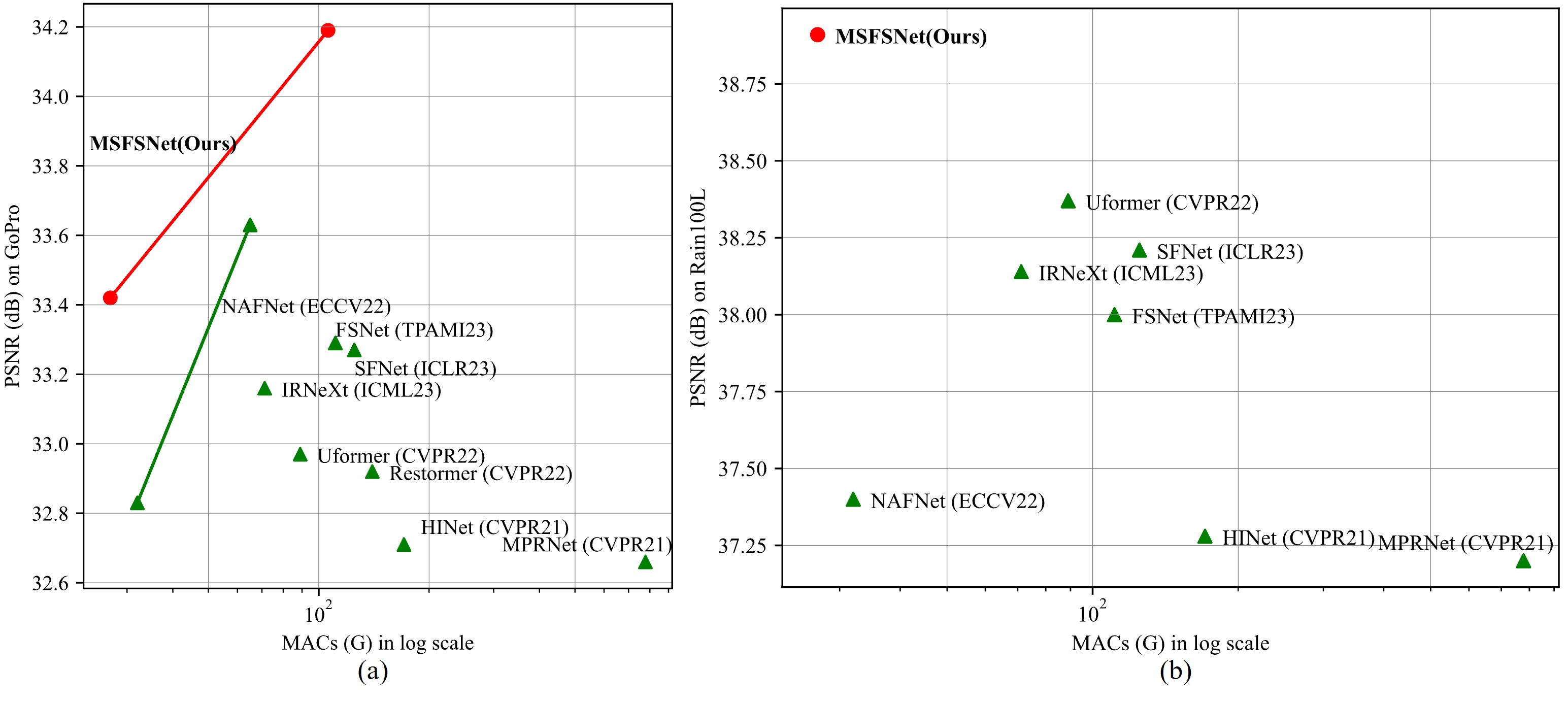}
    \caption{Computational cost vs. PSNR of models on Image Deblurring (Left) and Image Deraining (Right) tasks. Our MSFSNet achieve the SOTA performance with up to 69.7\% of cost reduction on Image Deraining. MACs are computed on patch size of 256 × 256.}
    \label{fig:param}
\end{figure*}

Finally, we implement a coarse-to-fine mode to alleviate the challenges in model training, incorporating multiple scales of input and output. It is noteworthy that our DFS and SFF serve as generic plug-in modules, seamlessly integrating into existing image restoration networks without requiring adjustments, resulting in performance improvements. 
As depicted in Figure~\ref{fig:param}, our MSFSNet model attains state-of-the-art performance while maintaining computational efficiency compared to existing methods.

The main contributions of this work are:
\begin{enumerate}
	\item We propose a multi-scale frequency selection network (MSFSNet) featuring two plug-in modules that can be seamlessly integrated into any existing restoration network.
    \item We design a dynamic filtering selection module (DFS) to generate high and low-frequency information using a learnable filter, discriminate which information to retain via the frequency cross-attention mechanism (FCAM), and select the most informative component for recovery.
    \item We develop a skip feature fusion block (SFF) to discriminatively determine the information that should propagate in skip connections via attention mechanism.
    \item Extensive experiments demonstrate that our MSFSNet achieves state-of-the-art performance across four typical image restoration tasks.
\end{enumerate}

\section{Related Work}
\subsection{Image Restoration}
Image restoration aims to enhance the quality of an image by addressing various degradations, which can be categorized into numerous sub-problems, including but not limited to image motion deblurring, defocus deblurring, deraining, and denoising. Due to its ill-posed nature, many conventional approaches~\cite{2013Unnatural,yang2020single, karaali2017edge, 2011Image, 2011Single} tackle this problem by relying on hand-crafted priors to constrain the set of plausible solutions. However, designing such priors is a challenging task and often lacks generalizability.

With the rapid advancement of deep learning, numerous works based on deep learning have gained significant popularity in the field of image restoration~\cite{chen2022simple,2021Rethinking,Zamir2021MPRNet, Chen_2021_CVPR,IDT,Zamir2021Restormer,Tsai2022Stripformer, Wang_2022_CVPR,mt10387581,ffanet,IRNeXt}, demonstrating more favorable performance compared to conventional methods. 
MPRNet~\cite{Zamir2021MPRNet} breaks down the overall recovery process into manageable steps, achieving a delicate balance between preserving spatial details and incorporating high-level contextualized information in image restoration.
FFANet~\cite{ffanet} utilizes an innovative attention mechanism and feature fusion structure, effectively preserving information from shallow layers into deep layers for direct haze-free image restoration.
MIRNet-V2~\cite{Zamir2022MIRNetv2} introduces a multi-scale architecture that maintains spatially-precise high-resolution representations throughout the network, gathering complementary contextual information from low-resolution representations.
NAFNet~\cite{chen2022simple} presents a streamlined baseline network for image restoration, achieved by either removing or replacing nonlinear activation functions. 
IRNeXt~\cite{IRNeXt} incorporates a multi-stage mechanism into a U-shaped network to enhance multi-scale representation learning. However, these methods rely on CNNs, and the intrinsic properties of convolutional operations, such as local receptive fields and fixed parameters, constrain the models' capability to efficiently eliminate long-range degradation perturbations and handle non-uniform degradation flexibly.

To overcome these limitations, Transformers~\cite{2017Attention} have been employed in image restoration~\cite{IPT,u2former,IDT,Zamir2021Restormer,Tsai2022Stripformer, Wang_2022_CVPR,mt10387581,liang2021swinir}, demonstrating superior performance compared to previous CNN-based baselines.
IPT~\cite{IPT} employs a Transformer-based multi-head multi-tail architecture, proposeing a pre-trained model for image restoration tasks. However, the computation of the scaled dot-product attention in Transformers leads to quadratic space and time complexity, rendering these methods computationally expensive for image restoration, especially considering the involvement of high-resolution inputs. In order to improve efficiency, Uformer~\cite{Wang_2022_CVPR}, SwinIR~\cite{liang2021swinir} and U$^2$former~\cite{u2former} computes self-attention based on a window. Restormer~\cite{Zamir2021Restormer} introduces transposed attention that implements self-attention across channels rather than the spatial dimension, resulting in the creation of an efficient transformer model. IDT~\cite{IDT} designs a  window-based and spatial-based
dual Transformer to achieve excellent results. 
Nevertheless, self-attention still relies on heavy computation.
In this paper, rather than exploring variants of CNNs and Transformers, we focus on designing plug-and-play component modules that can be incorporated into existing image restoration networks without any adjustments. These modules assist in capturing multi-scale frequency information and selecting the most informative components for recovery, resulting in performance improvements.

\subsection{Frequency Learning}
In recent years, some deep learning-based vision research has started exploring the frequency domain due to the ease of handling different frequency subbands.~\cite{Chi_2020_FFC} proposes a generic fast Fourier convolution (FFC) operator to overcome the limitations of convolutional local receptive fields and achieves good performance on high-level vision tasks. Motivated by the great success of high-level vision tasks, frequency learning have
been applied to image restoration~\cite{kong2023efficient,fLi2023ICLR,f8803391,fxint2023freqsel,WACAFRN,huang2022WINNet,mf9786841,mf9917526,FSNet, SFNet, 10196308}. 
WINNet~\cite{huang2022WINNet} proposes a wavelet-inspired invertible network that combines the strengths of wavelet-based and learning-based approaches, offering high interpretability and strong generalization ability.
Leveraging the convolution theorem, which states that convolution in the spatial domain equals point-wise multiplication in the frequency domain, FFTformer~\cite{kong2023efficient} develops an effective and efficient method that explores the properties of Transformers for high-quality image deblurring. 
UHDFour~\cite{fLi2023ICLR} incorporates Fourier into the network, processing the amplitude and phase of a low-light image separately to prevent noise amplification during luminance enhancement.
DeepRFT~\cite{fxint2023freqsel} integrates Fourier transform to incorporate kernel-level information into image deblurring networks.
WACAFRN~\cite{WACAFRN} proposes a fine-grained residual network for image denoising, combining an adaptive coordinate attention mechanism with cascaded Res2Net residual blocks in the encoder and incorporating global and local residual blocks guided by wavelet and adaptive coordinate attention in the decoder.
JWSGN~\cite{mf9786841} utilizes wavelet transform to separate different frequency information in the image and then reconstructs this information using a multi-branch network, aiming to enhance the effectiveness in reconstructing high-frequency details.
FGDNet~\cite{mf9917526} proposes a frequency-domain guided denoising algorithm, utilizing a well-aligned guidance image to enhance denoising while considering both the representation model of the guidance image and fidelity to the noisy target.

Despite their merits, the mentioned methods neglect the effective utilization of both low-frequency and high-frequency information. To address these limitations and choose the most informative frequency component for reconstruction, AirFormer~\cite{10196308}  propose
a supplementary prior  module  to selectively filter high-frequency components, ensuring the preservation of low-frequency components. SFNet~\cite{SFNet} and FSNet~\cite{FSNet} utilize a multi-branch and content-aware module to dynamically and locally decompose features into separate frequency subbands, selecting the most informative components for recovery.
However, these methods only fuses low-frequency and high-frequency information in the channel dimension, often face challenges in distinguishing background details in local regions, and cannot strike a balance between overlapped scenes. 
In this paper, we present two plug-in modules: (1) the dynamic filter selection module (DFS) to select the most informative frequency  to recover, and (2) the skip feature fusion block (SFF) to discriminatively determine the information that should propagate in skip connections. DFS and SFF can be seamlessly integrated into existing networks without requiring any adjustments, resulting in performance improvements.

\section{Method}
In this section, we first outline the overall pipeline of our MSFSNet. Subsequently, we delve into the details of the proposed dynamic filter selection module (DFS)  within  multi-scale frequency selention block (MSFS), and skip feature fusion block (SFF).

\subsection{Overall Pipeline} 
\begin{figure*}[htb] 
	\centering
	\includegraphics[width=1\textwidth]{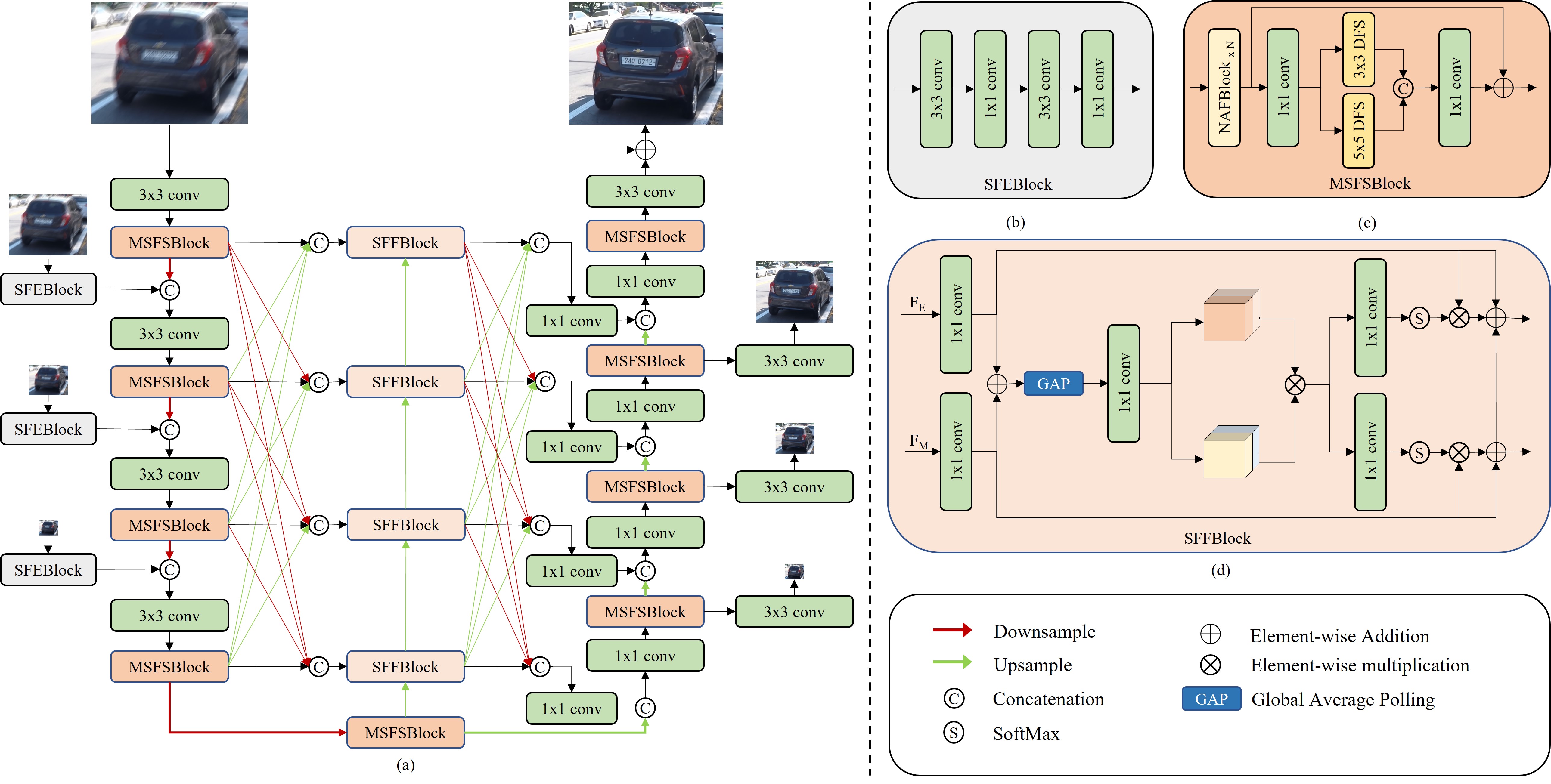}
	\caption{
(a) Overall architecture of MSFSNet. (b) Shallow feature extraction block (SFE) for capturing shallow features in low-resolution input images. (c) Multi-scale frequency selection block (MSFS) integrating spatial and frequency domain knowledge, and select the most informative component for recovery. (d) Skip feature fusion block (SFF) for discriminative information propagation in skip-connections.}
	\label{fig:network}
\end{figure*}

The overall pipeline of our proposed MSFSNet, shown in Figure~\ref{fig:network}, is based on a  hierarchical encoder-decoder. 

Given a degraded image $\mathbf{I} \in \mathbb R^{H \times W \times 3}$, 
MSFSNet initially applies a convolution to acquire shallow features $\mathbf{F_{0}} \in \mathbb R^{H \times W \times C}$  ($H, W, C$ are the feature map height, width, and channel number, respectively). 
These shallow features undergo a four-scale encoder sub-network, progressively decreasing resolution while expanding channels. It's essential to note the use of multi-input and multi-output mechanisms for improved training. The low-resolution degraded images are incorporated into the main path through the the SFE and concatenation, followed by 3 × 3 convolution to adjust channels.
The in-depth features then enter a middle block, and the resulting deepest features feed into a four-scale decoder, gradually restoring features to the original size. During this process, the encoder features, determined through SFF discrimination, are concatenated with the decoder features to facilitate the reconstruction.
Finally, we apply  convolution to the refined features to generate residual image $\mathbf{X}\in \mathbb R^{H \times W \times 3}$ to which degraded image is added to obtain the restored image: $\mathbf{\hat{I}} = \mathbf{X} +\mathbf{I}$.  It's important to note that the three low-resolution results are solely used for training. To facilitate the frequency selection process, we optimize the proposed network adopt $L_1$ loss in both spatial and frequency domains: 
\begin{equation}
\label{eq:loss}
\begin{aligned}
     L_{s} &= \sum_{i=1}^{4} \frac{||\mathbf{\hat{I_i}}-\mathbf{\overline I_i}||_1}{4}
   \\
    L_{f} &= \sum_{i=1}^{4} \frac{||\mathcal{F}(\mathbf{\hat{I}_i})-\mathcal{F}(\mathbf{\overline I_i})||_1}{4}
    \\
    L &= L_s + \lambda L_f 
\end{aligned}
\end{equation}
where $i$ denotes the index of input/output images at different scales, $\mathcal{F}$ represents fast Fourier transform, $\mathbf{\overline I_i}$ denotes the target images, and the parameter $\lambda$ controls the relative importance of the two loss terms, set to 0.1 as in~\cite{FSNet}.

\subsection{Multi-scale Frequency Selection Block (MSFS)}
Many existing image restoration methods have achieved good results, but they mainly concentrate on the spatial domain, neglecting the understanding of frequency variation. While some methods utilize transformation tools to explore the frequency domain, they often lack the flexibility to selectively choose the most informative frequencies and may require corresponding reverse transformation operations, leading to increased computational complexity. 
To address this limitation, we introduce the multi-scale frequency selection block (MSFS), which captures spatial features by NAFBlock~\cite{chen2022simple} and feeds them into two parallel dynamic filter selection modules (DFS) at different scales to integrate frequency knowledge. Formally, given the input features $X$, the procedures of MSFS can be defined as:
\begin{equation}
\label{eq:msfs}
\begin{aligned}
    X_s &= X \oplus NAF(X)
    \\
    X_f^3 &= DFS^3(f_{1 \times 1}^c(X_s))
    \\
    X_f^5 &= DFS^5(f_{1 \times 1}^c(X_s))
    \\
    X_o &= X_s \oplus f_{1 \times 1}^c([X_f^3, X_f^5])
\end{aligned}
\end{equation}
where $X_s$, $X_f$ and $X_o$ denote the spatial features, frequency features and the output of MSFS; $f_{1 \times 1}^c$ represents $1 \times 1$ convolution; $[\cdot]$ represents the channel-wise concatenation. Next, we will describe NAFBlock and DFS below.

\begin{figure}[htb] 
	\centering
	\includegraphics[width=0.5\textwidth]{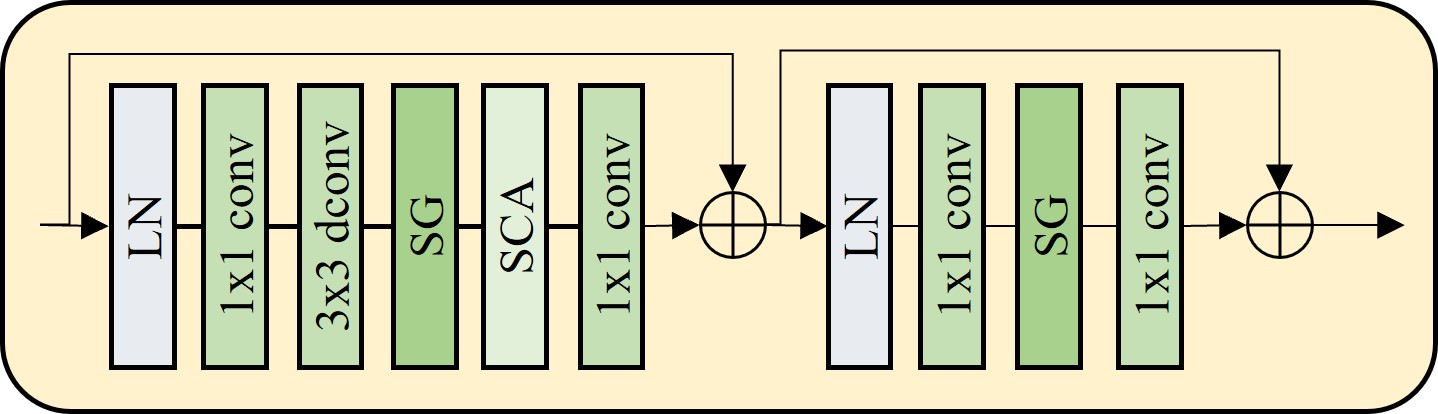}
	\caption{The structure of NAFBlock~\cite{chen2022simple}. }
	\label{fig:nafnet}
\end{figure}

\subsubsection{\textbf{NAFBlock}}
In this paper, our primary focus is on designing two plug-and-play modules DFS and SFF. To assess the effectiveness of these modules, we utilize NAFBlock~\cite{chen2022simple} as our backbone for extracting spatial features (See Figure~\ref{fig:nafnet}). Specifically, given an input tensor $X_{l-1}$, we initially process it  through Layer Normalization (LN), Convolution, Simple Gate (SG), and Simplified Channel Attention (SCA) to obtain spatial  features $X^{s}_{l}$ as follows:
\begin{equation}
\begin{aligned}
	\label{equ:0msb}
	X^{'}_{l-1} &= SCA(SG(f_{3 \times 3}^{dwc} (f_{1 \times 1}^c(LN(X_{l-1}))))
\\
	X^{s}_{l-1} &= f_{1 \times 1}^c(X^{'}_{l-1}) \oplus X_{l-1}
 \\
 X^s_{l} &= X^{s}_{l-1} \oplus f_{1 \times 1}^c(SG(f_{1 \times 1}^c(LN(X^{s}_{l-1}))))
\\
    SCA(X_{f3}) &= X_{f3} \otimes f_{1 \times 1}^c( GAP(X_{f3}))
\\
    SG(X_{f0}) &= X_{f1} \otimes X_{f2} 
\end{aligned}
\end{equation}
where  $f_{3 \times 3}^{dwc}$ denotes the $3 \times 3$ depth-wise convolution, and GAP is the global average pooling. $X_{f1}, X_{f2} \in \  R^{H \times W \times \frac{C}{2}}$  is obtained by splits $X_{f0}$ along channel dimension.

\begin{figure*}[htb] 
	\centering
	\includegraphics[width=1\textwidth]{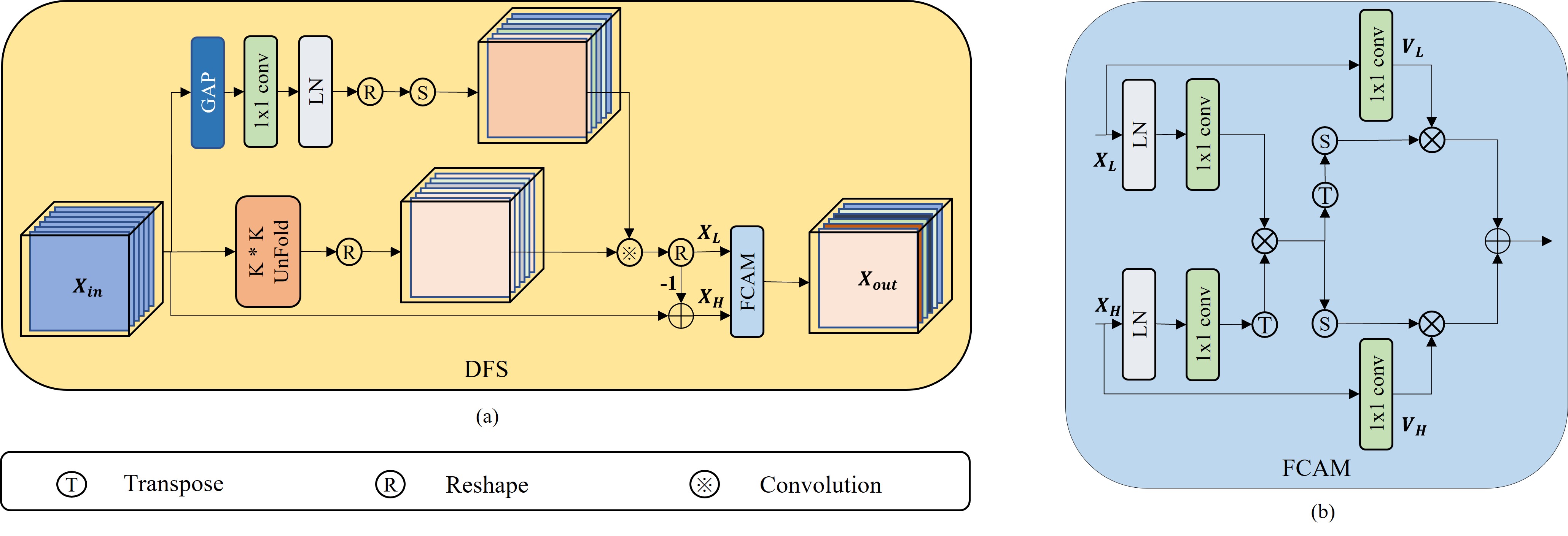}
	\caption{(a) The structure of  dynamic filter selection modules (DFS). (b) Frequency cross-attention mechanism (FCAM) to discriminate which low-frequency and high-frequency information should be retained.}
	\label{fig:dfs}
\end{figure*}

\subsubsection{\textbf{Dynamic Filter Selection Module (DFS)}}

To integrate frequency knowledge and automatically select the most informative components for recovery, we introduce dynamic filtering selection modules (DFS). As shown in Figure~\ref{fig:dfs} (a), DFS comprises two key processes: (1) Dynamically generating high and low-frequency maps using a learnable filter. (2) Employing the frequency cross-attention mechanism (FCAM) to discriminate which low-frequency and high-frequency information should be retained.

To dynamically decompose the feature map, we draw inspiration from~\cite{FSNet,SFNet,IRNeXt} and utilize a learnable low-pass filter shared across group dimensions to generate the low-frequency map. The high-frequency map is then obtained by subtracting the low-frequency map from the original feature map. Specifically, given an input feature $F \in \mathbb R^{H \times W \times C}$, we first use the GAP, convolution, LN, reshape and softmax activation function to produce a low-pass filter for each group of the input as follows:
\begin{equation}
\label{eq:fl}
F_l = Softmax(Reshape(LN( f_{1 \times 1}^c(GAP(F)))))
\end{equation}
where he reshape operation divides the feature map from $\mathbb R^{1 \times 1 \times gk^2}$ to
$\mathbb R^{g \times k \times k}$, $g$ denotes the number
of groups, $k \times k$ is the kernel size of low-pass filters, controlling the scale in DFS. In this paper, we use the $3 \times 3$ and $5 \times 5$ shown in Figure~\ref{fig:network} (c). Then we employ  the $k \times k$ kernel sizes to unfold the input feature  $F$ and reshape it. The resulting outcome is convolved with the low-pass filter $F_l$, and finally reshaped back to the original feature map size to obtain the final low-frequency feature map $X_L$. The process can be expressed as:
\begin{equation}
	\label{equ:fl}
	X_L =  F_l \ocoasterisk Reshape(UnFold(F))
\end{equation}

To obtain the high-frequency feature map $X_H$, we subtract the resulting low-frequency feature map $X_L$  from the input feature $F$, which is expressed as:
\begin{equation}
	\label{equ:fh}
	X_H =  F - X_L
\end{equation}

We revisit all the previously proposed modulator modules~\cite{SFNet, FSNet} that utilize channel-wise attention to emphasize the truly useful components for reconstruction. And we design a frequency cross-attention mechanism (FCAM) to determine which low-frequency and high-frequency information should be retained. As illustrated in Figure~\ref{fig:dfs} (b), we compute the dot products between the query $Q\in \mathbb R^{H \times W \times C}$ projected by the source frequency feature (e.g., low-frequency), and the key, value $K,V \in \mathbb R^{H \times W \times C}$ projected using the target frequency feature (e.g., high-frequency). This is followed by applying a softmax function to derive the weights assigned to the values:
\begin{equation}
	\label{equ:102}
    Attention(Q, K, V) =  Softmax(\frac{QK^T}{\beta})V
\end{equation}
where $\beta$ is a learning scaling parameter used to adjust the magnitude of the dot product of $Q$ and $K$ prior to the application of the softmax function defined by $\beta = \sqrt{C}$. 

Specifically, given a pair of low- and high-frequency features $X_L, X_H \in \mathbb R^{H \times W \times C}$, we begin by applying layer normalization, and subsequently derive the feature $\hat{X_L}$ and $\hat{X_H}$ using a $1 \times 1$ convolution layer.
Noted that, $Q_L = X_L^{'}$, $K_L = X_H^{'}$ and $Q_R = X_R^{'}$, $K_R = X_L^{'}$. Next, we generate the value matrix $V_L$ and $V_H$ by using a $1 \times 1$ convolution layer, respectively. Subsequently, we compute bidirectional cross-attention between the low- and high-frequency features as follows:
\begin{equation}
    \begin{aligned}
        F_{H \rightarrow L} &= Attention(Q_H, K_H , V_H)
        \\
        F_{L \rightarrow H} &= Attention(Q_L, K_L, V_L)
    \end{aligned}
\end{equation}

Finally, we fuse the interacted cross-frequency information $F_{H \rightarrow L}$ and $ F_{L \rightarrow H}$ by element-wise addition to obtain the most informative frequency $F_o$, 
\begin{equation}
\label{eq:rcam}
        F_o = \lambda_L F_{L \rightarrow H} \oplus \lambda_H F_{H \rightarrow L}
\end{equation}
where $\lambda_L$ and $\lambda_H$ are channel-wise scale parameters that are trainable and initialized with zeros to aid in stabilizing training.

\subsection{Skip Feature Fusion Block (SFF)}
To improve the image restoration capabilities of the model, it is common to establish skip connections between the encoder and decoder features through addition or concatenation. However, the features from the encoder may include image degradation factors, and the simple addition or concatenation process for feature interaction and aggregation between the encoder and decoder is prone to implicit noise, thereby affecting the model's image restoration ability. 
To address this issue, as shown in Figure~\ref{fig:network} (d), we introduce a skip feature fusion block (SFF) that utilizes contextual features to selectively determine which information should be propagated through skip connections. 
To elaborate further, for the SFFBlock at the lowest level, we initially concatenate the  encoder features $XE_i, (i=1,2,3,4)$  and adjust the number of channels through a convolution:
\begin{equation}
\label{eq:sff1}
        FE_1 = f_{1 \times 1}^c([XE_1, XE_2, XE_3, XE_4])
\end{equation}

Subsequently, we obtain $FM_1$ by upsampling the feature map of the middle block, and use it to selectively re-weight the features $FE_1$ through a self-attention mechanism, obtaining the feature $\hat{FE_1}$ that preserves the most useful information, the process is as follows:
\begin{equation}
\begin{aligned}
\label{equ:sff2}
    FS_1 &= SG(f_{1 \times 1}^c(GAP(FE_1+FM_1)))
    \\
    FE^m_1 &= Softmax(f_{1 \times 1}^c(FS_1))
    \\
    FM^m_1 &= Softmax(f_{1 \times 1}^c(FS_1))
      \\
    \hat{FM_1} &= FM_1 \otimes FM^m_1 \oplus FM_1
      \\
    \hat{FE_1} &=  FE_1 \otimes FE^m_1 \oplus FE_1 \oplus \hat{FM_1}
\end{aligned}
\end{equation}

Through the aforementioned process, we obtain the encoder features $\hat{FE_i}, (i=1,2,3,4)$ that retain useful information for all levels. Finally, the features ($\hat{FE_i}$) are concatenated and a convolution operation is applied to adjust the number of channels, resulting in the final feature which will fuse with the decoder.  
SFF offers two pivotal advantages. First, it selectively identifies the information to be transmitted in the skip connection, mitigating the interference of implicit noise. Second, it combines features at different scales while preserving their distinctive complementary properties. Consequently, the model is capable of acquiring a diverse and accurate feature set.

\begin{figure*}[htb] 
	\centering
	\includegraphics[width=1\textwidth]{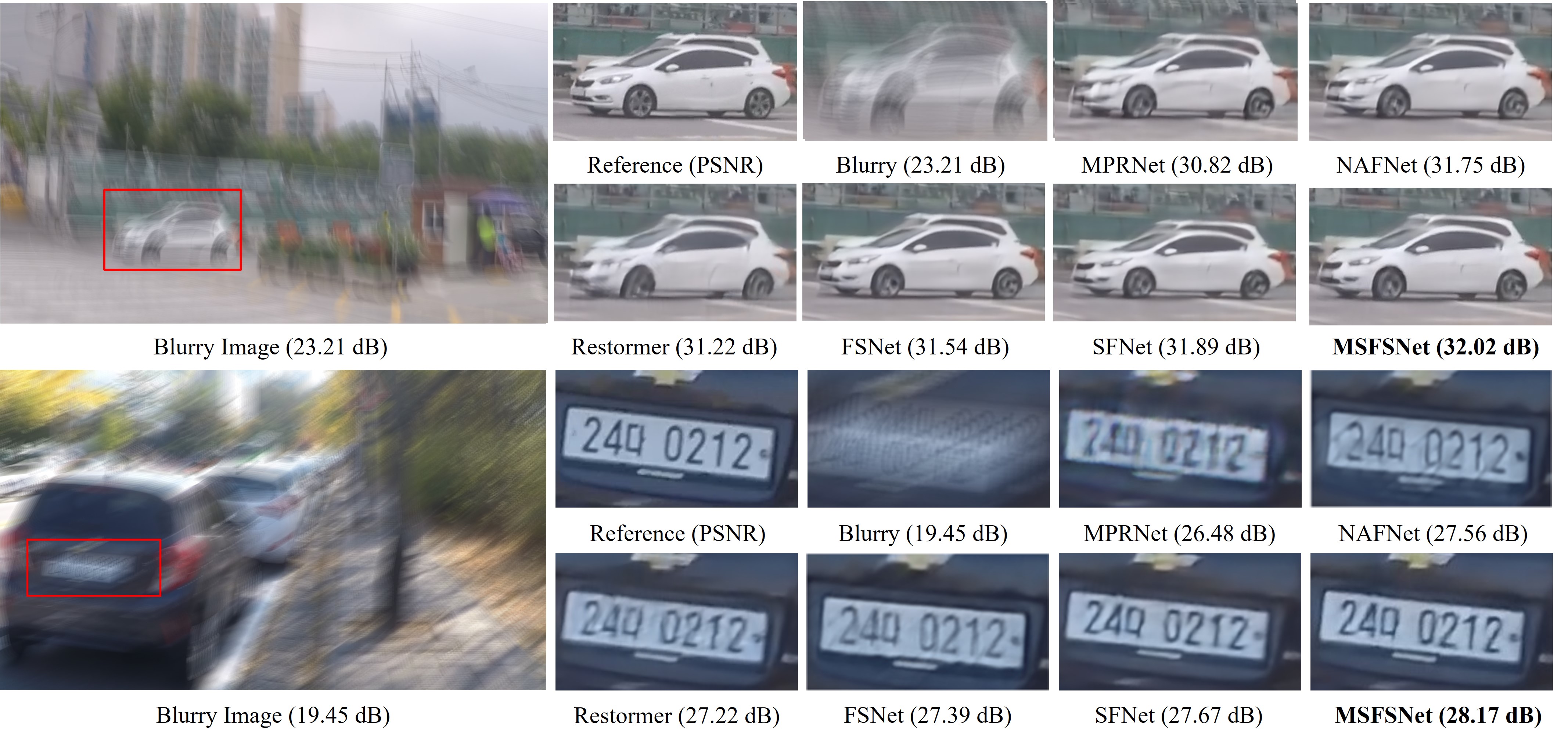}
	\caption{\textbf{Image motion deblurring} comparisons on the GoPro dataset~\cite{Gopro}. Compared to the state-of-the-art methods, our MSFSNet excels in restoring sharper and perceptually faithful images. }
	\label{fig:blurm}
\end{figure*}

\section{Experiments}
In this section, we provide details about the experimental settings and then present both qualitative and quantitative comparisons between MSFSNet and other state-of-the-art methods. Following that, we conduct ablation studies to validate the effectiveness of our approach. Finally, we assess the resource efficiency of MSFSNet.

\subsection{Experimental Settings}
In this section, we introduce the details of the used datasets, and training details.

\subsubsection{\textbf{Datasets}}
In this section, we will introduce the datasets used and provide details regarding the training configurations.

\textbf{Image Motion Deblurring.} 
Following recent methods~\cite{FSNet,Zamir2021MPRNet}, we train MSFSNet using the GoPro dataset~\cite{Gopro}, which includes 2,103 image pairs for training and 1,111 pairs for evaluation. To assess the generalizability of our approach, we directly apply the GoPro-trained model to the test images of the HIDE dataset~\cite{HIDE}, which consists of 2,025 images for evaluation. Both the GoPro~\cite{Gopro} and HIDE~\cite{HIDE} datasets are synthetically generated, whereas the RealBlur dataset~\cite{realblurrim_2020_ECCV} contains image pairs captured under real-world conditions, divided into two subsets: RealBlur-J and RealBlur-R.

\textbf{Single-Image Defocus Deblurring.}
To assess the effectiveness of our method, we utilize the DPDD dataset~\cite{DPDNet}, following the approach of recent methods~\cite{FSNet,Zamir2021Restormer}. This dataset comprises images from 500 indoor/outdoor scenes captured with a DSLR camera. Each scene includes three defocused input images and a corresponding all-in-focus ground-truth image, labeled as right view, left view, center view, and the all-in-focus ground truth. The DPDD dataset is divided into training, validation, and testing sets with 350, 74, and 76 scenes, respectively. MSFSNet is trained using the center view images as input, computing loss values between outputs and corresponding ground-truth images.

\textbf{Image Deraining.}
Following the experimental setups of recent state-of-the-art methods for image deraining~\cite{Zamir2021MPRNet,FSNet}, we train our model using 13,712 clean-rain image pairs collected from multiple datasets~\cite{Rain100,Test100,8099669,7780668}. Using the trained MSFSNet, we conduct evaluations on various test sets, including Rain100H~\cite{Rain100}, Rain100L~\cite{Rain100}, Test100~\cite{Test100}, and Test1200~\cite{DIDMDN}.

\textbf{Image Denoising.}
In line with the methodology of~\cite{Zamir2021Restormer}, we train a single MSFSNet model capable of handling various noise levels on a composite dataset. This dataset comprises 800 images from DIV2K~\cite{DIK}, 2,650 images from Flickr2K~\cite{lim2017enhanced}, 400 images from BSD500~\cite{BSD500}, and 4,744 images from WED~\cite{ma2016waterloo}. The noisy images are generated by adding additive white Gaussian noise with a random noise level $\sigma$ chosen from the set [15, 25, 50] to the clean images. Testing is conducted on the CBSD68~\cite{BSD68}, Urban100~\cite{urban100}, and Kodak24~\cite{kodak} benchmark datasets.

\begin{table}[ht]
\centering
\caption{Image motion deblurring results.  The best and second best scores are \textbf{highlighted} and \underline{underlined}. Our MSFSNet-B and MSFSNet  are trained only on the GoPro dataset but achieves the highest and second-highest scores on the average of the effects on both datasets, respectively. \label{tb:deblur}}
\resizebox{\linewidth}{!}{
\begin{tabular}{ccccc||cc}
    \hline
    \multicolumn{1}{c}{} & \multicolumn{2}{c}{GoPro~\cite{Gopro}}  & \multicolumn{2}{c||}{HIDE~\cite{HIDE}} & \multicolumn{2}{c}{Average}
    \\
   Methods & PSNR $\uparrow$ & SSIM $\uparrow$ & PSNR $\uparrow$ & SSIM $\uparrow$   &  PSNR $\uparrow$ &  SSIM $\uparrow$
    \\
    \hline\hline

    SPAIR~\cite{SPAIR} & 32.06 & 0.953 & 30.29 & 0.931 &31.18  &0.942
    \\
    MIMO-UNet++~\cite{2021Rethinking} & 32.45 & 0.957 & 29.99 & 0.930 &31.22 &0.944
    \\
    MPRNet~\cite{Zamir2021MPRNet} & 32.66 & 0.959 & 30.96 & 0.939 &31.81  &0.949
    \\
    MPRNet-local~\cite{Zamir2021MPRNet} & 33.31 & 0.964 &31.19 &0.945 &32.25 &0.955
    \\
    HINet~\cite{Chen_2021_CVPR}&32.71&0.959&30.32&0.932&31.52&0.946
    \\
    HINet-local~\cite{Chen_2021_CVPR}&33.08&0.962&-&-&-&-
    \\
    Uformer~\cite{Wang_2022_CVPR} &32.97 & \underline{0.967} &30.83 &\underline{0.952} &31.90  &\underline{0.960}
     \\
    MSFS-Net~\cite{MSFSnet} & 32.73 & 0.959 & 31.05 & 0.941& 31.99 & 0.950 
    \\
    MSFS-Net-local~\cite{MSFSnet} & 33.46 & 0.964 & 31.30 & 0.943 & 32.38 & 0.954 
    \\
    NAFNet-32~\cite{chen2022simple}&32.83&0.960&-&-&-&-
    \\
    NAFNet-64~\cite{chen2022simple}&\underline{33.62}&\underline{0.967}&-&-&-&-
    \\
    Restormer~\cite{Zamir2021Restormer} & 32.92 & 0.961 & 31.22 & 0.942 &32.07 &0.952
    \\
    Restormer-local~\cite{Zamir2021Restormer} & 33.57 & 0.966 & 31.49 & 0.945 &32.53  &0.956
    \\
    IRNeXt~\cite{IRNeXt} &33.16 &0.962 &- & - &- &-
    \\
    SFNet~\cite{SFNet} &33.27 &0.963 &31.10 & 0.941 &32.19 &0.952
    \\
    FSNet~\cite{FSNet} &33.29&0.963 &31.05 & 0.941 &32.17 &0.952
    \\
    \hline
    \textbf{MSFSNet(Ours)} &33.42&0.964&\underline{31.68}&0.947 &\underline{32.55} &0.956
    \\
    \textbf{MSFSNet-B(Ours)} &\textbf{34.19}&\textbf{0.969}&\textbf{32.67}&\textbf{0.954} &\textbf{33.43} &\textbf{0.962}
    \\
    \hline
\end{tabular}}
\end{table}

\begin{table}
\centering
\caption{Quantitative real-world deblurring results.}
\label{tb:0deblurringreal}
\begin{tabular}{ccccc}
    \hline
    \multicolumn{1}{c}{} & \multicolumn{2}{c}{RealBlur-R}  & \multicolumn{2}{c}{RealBlur-J} 
    \\
   Methods & PSNR $\uparrow$ & SSIM $\uparrow$ & PSNR $\uparrow$ & SSIM $\uparrow$   
    \\
    \hline
DeblurGAN-v2~\cite{deganv2} & 36.44 & 0.935& 29.69& 0.870
\\
    MPRNet~\cite{Zamir2021MPRNet} & 39.31 & 0.972 & 31.76 & 0.922
   \\
Stripformer~\cite{Tsai2022Stripformer} & 39.84 & 0.975 & 32.48 & 0.929
\\
FFTformer~\cite{kong2023efficient}&40.11& 0.973 &32.62 &0.932
\\
    \textbf{MSFSNet(Ours)}&\textbf{40.62}&\textbf{0.977}&\textbf{32.85}&\textbf{0.936}
    \\
    \hline
\end{tabular}
\end{table}

\begin{table*}[hb]
    \centering
        \caption{Quantitative comparisons with previous leading single-image defocus deblurring methods on the DPDD testset~\cite{DPDNet} (containing 37 indoor and 39 outdoor scenes).}
    \label{tab:deblurd}
    \resizebox{\linewidth}{!}{
    \begin{tabular}{c|ccc|ccc|ccc}
        \hline
    \multicolumn{1}{c|}{} & \multicolumn{3}{c|}{Indoor Scenes}  & \multicolumn{3}{c|}{Outdoor Scenes} & \multicolumn{3}{c}{Combined}
    \\
   Methods & PSNR $\uparrow$ & SSIM $\uparrow$ &MAE $\downarrow$  & PSNR $\uparrow$ & SSIM $\uparrow$  &MAE $\downarrow$    &  PSNR $\uparrow$ &  SSIM $\uparrow$ &MAE $\downarrow$  
    \\
    \hline
    \hline
JNB~\cite{JNB} &26.73 &0.828 &0.031  &21.10 &0.608 &0.064  &23.84 &0.715 &0.048 
\\
DPDNet~\cite{DPDNet} & 26.54 &0.816 &0.031  &22.25 &0.682 &0.056  &24.34 &0.747 &0.044
\\
KPAC~\cite{KPAC}& 27.97 &0.852 &0.026  &22.62 &0.701 &0.053  &25.22 &0.774 &0.040 
\\
IFAN~\cite{IFAN}& 28.11 &0.861 &0.026  &22.76 &0.720 &0.052  &25.37 &0.789  &0.217
\\
Restormer~\cite{Zamir2021Restormer}& 28.87 &\underline{0.882} &0.025 &23.24 &0.743 &0.050  &25.98 &0.811 &0.038 
\\
IRNeXt~\cite{IRNeXt} &\underline{29.22} &0.879 &0.024  &\underline{23.53} &\underline{0.752} &\underline{0.049}  &\underline{26.30} &\underline{0.814} &\underline{0.037} 
\\
SFNet~\cite{SFNet} &29.16 &0.878 &\underline{0.023}  &23.45 &0.747 &\underline{0.049}  &26.23 &0.811 &\underline{0.037} 
\\
FSNet~\cite{FSNet} &29.14 &0.878 &0.024  &23.45 &0.747 &0.050  &26.22 &0.811 &\underline{0.037} 
\\
\hline
\textbf{MSFSNet(Ours)}&\textbf{29.25}	&\textbf{0.929}	&\textbf{0.022}		&\textbf{23.84}	&\textbf{0.802}	&\textbf{0.044}		&\textbf{26.47}	&\textbf{0.859}	&\textbf{0.033}	

    \\
    \hline
    \end{tabular}}
\end{table*}

\subsubsection{\textbf{Training details}}
For different tasks, we train separate models, and unless specified otherwise, the following parameters are employed. The models are trained using the Adam optimizer~\cite{2014Adam} with parameters $\beta_1=0.9$ and $\beta_2=0.999$. The initial learning rate is set to $2 \times 10^{-4}$ and gradually reduced to $1 \times 10^{-7}$ using the cosine annealing strategy~\cite{2016SGDR}. The batch size is chosen as $32$, and patches of size $256 \times 256$ are extracted from training images. Data augmentation involves horizontal and vertical flips. We scale the network width by setting the number of channels to 32 and 64 for MSFSNet and MSFSNet-B, respectively.  For each scale of  encoder we use [1,1,1,28] and  [1,1,1,1] for  decoder.

\subsection{Experimental Results}
\subsubsection{ \textbf{Image Motion Deblurring}}

The quantitative results on the GoPro and HIDE dataset are shown in Table~\ref{tb:deblur}. Our MSFSNet-B and MSFSNet obtain generates a  gain of 0.02 dB and 0.9 dB PSNR  over Restormer-Local~\cite{Zamir2021Restormer} on the average of the effects on both datasets, respectively. 
In particular, our MSFSNet-B shows 0.57 dB performance improvement over NAFNet-64~\cite{chen2022simple} on GoPro~\cite{Gopro}. Moreover, as shown in Figure~\ref{fig:param}, through the scaling up of the model size, our MSFSNet achieves even better performance, highlighting the scalability of MSFSNet. Noted that, even though our network is trained solely on the GoPro~\cite{Gopro} dataset, it still achieves a substantial gain of 1.18 dB PSNR over Restormer-Local~\cite{Zamir2021Restormer} on the HIDE~\cite{HIDE} dataset. This demonstrates its impressive generalization capability. Figure.~\ref{fig:blurm} shows our model recovered more visually pleasant results.

\begin{figure}[htb] 
	\centering
	\includegraphics[width=1\linewidth]{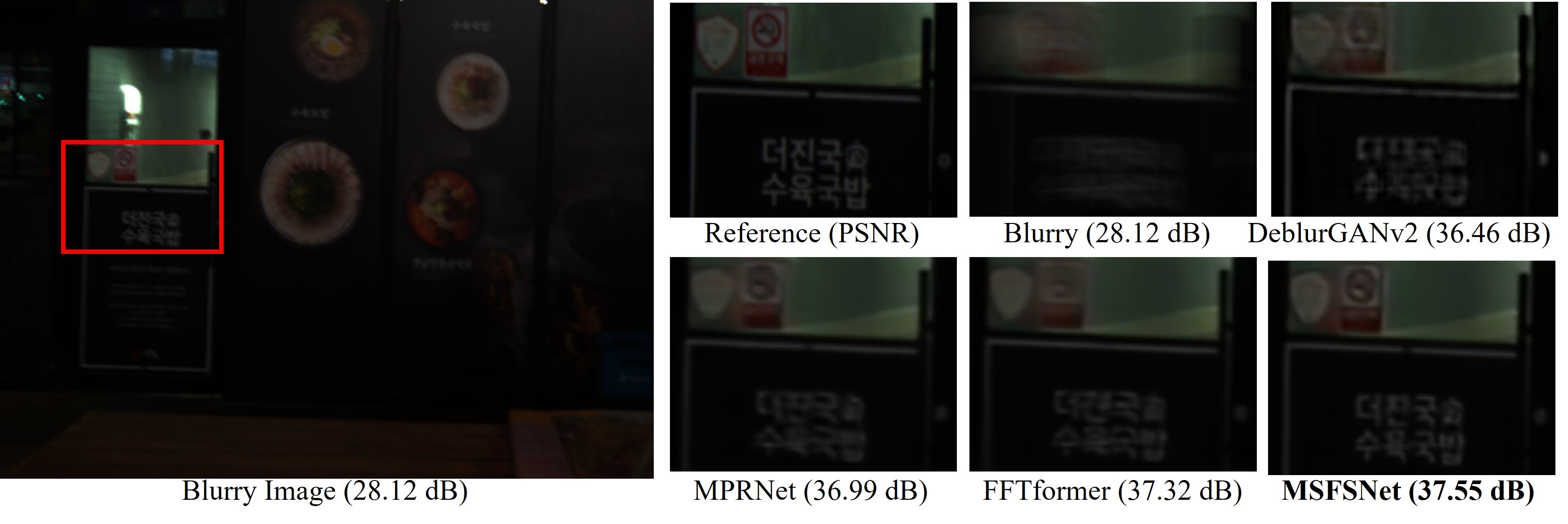}
	\caption{\textbf{Comparison of image motion deblurring} on the RealBlur dataset~\cite{realblurrim_2020_ECCV}.}
	\label{fig:real2}
\end{figure}

Additionally, for real-world scenes, we evaluate the performance of our MSFSNet using real-world images from the RealBlur dataset~\cite{realblurrim_2020_ECCV}. The results are presented in Table~\ref{tb:0deblurringreal}, our MSFSNet achieves performance gains of 0.51 dB on the RealBlur-R subset over FFTformer~\cite{kong2023efficient} and 0.23 dB on the RealBlur-J subset.
Figure~\ref{fig:real2} presents visual comparisons of the evaluated approaches. Overall, the images restored by our model exhibit sharper details.

\subsubsection{\textbf{Single-Image Defocus Deblurring}}
\begin{figure*}[htb] 
	\centering
	\includegraphics[width=1\textwidth]{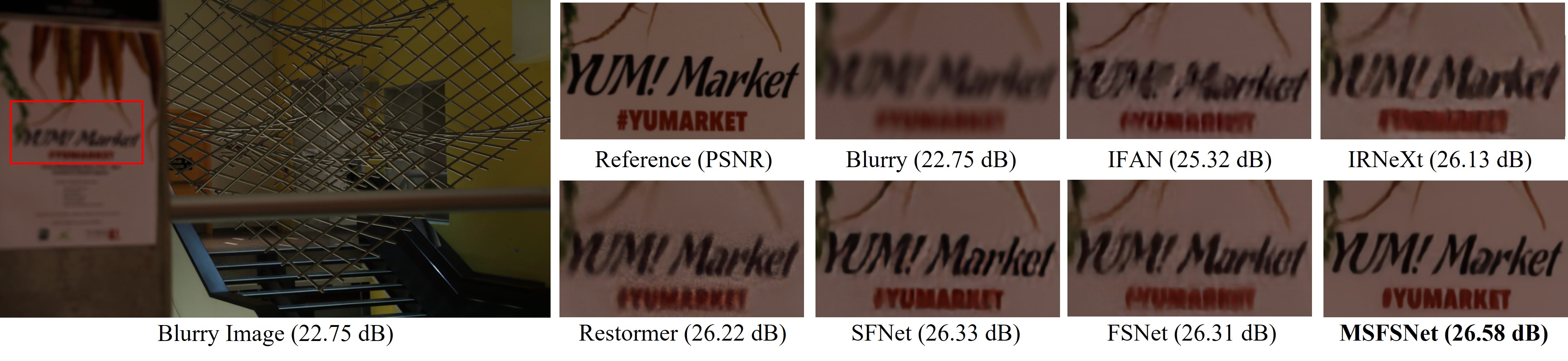}
	\caption{\textbf{Single image defocus deblurring} comparisons on the DDPD dataset~\cite{DPDNet}. Compared to the state-of-the-art methods, our MSFSNet effectively removes blur while preserving the fine image details. }
	\label{fig:blurd}
\end{figure*}

Table~\ref{tab:deblurd} presents a numerical comparison of defocus deblurring methods on the DPDD dataset~\cite{DPDNet}. MSFSNet outperforms other state-of-the-art methods across most metrics. Notably, in the outdoor scenes category, MSFSNet exhibits a 0.31 dB improvement over the leading method IRNeXt~\cite{IRNeXt}. Additionally, our results demonstrate significant improvement in the SSIM metric, indicating better preservation of structural similarity. The visual results in Figure~\ref{fig:blurd} illustrate that our method recovers more details and visually aligns more closely with the ground truth compared to other algorithms.

\begin{table*}
\centering
\caption{Image deraining results. When averaged across all four datasets, our MSFSNet is better than the state-of-the-art by 0.78 dB. }\label{tb:derain}
\resizebox{\linewidth}{!}{
\begin{tabular}{ccccccccc||cc}
    \hline
    \multicolumn{1}{c}{} & \multicolumn{2}{c}{Test100~\cite{Test100}}  & \multicolumn{2}{c}{Test1200~\cite{DIDMDN}} & \multicolumn{2}{c}{Rain100H~\cite{Rain100}} & \multicolumn{2}{c||}{Rain100L~\cite{Rain100}} & \multicolumn{2}{c}{Average} 
    \\
   Methods &PSNR $\uparrow$ &  SSIM $\uparrow$  & PSNR $\uparrow$ & SSIM $\uparrow$ &PSNR $\uparrow$ &SSIM $\uparrow$ & PSNR $\uparrow$&SSIM $\uparrow$ &PSNR $\uparrow$ & SSIM $\uparrow$
    \\
    \hline\hline
    
     MPRNet~\cite{Zamir2021MPRNet}  & 30.27 & 0.907 & 32.91 &  0.916   & 30.51 & 0.890  & 37.20 & 0.965 & 32.73 & 0.921
       \\
     SPAIR~\cite{SPAIR}  & 30.35 & 0.909 & 33.04 &  0.922   & 30.95 & 0.893  & 37.30 & 0.978& 32.91 & 0.926
     \\
    
   
     Restormer~\cite{Zamir2021Restormer} &\underline{32.00} & \textbf{0.923} & \underline{33.19} & \underline{0.926} & 31.46 &0.904 &\underline{38.99} &0.978 &\underline{33.91} & \underline{0.933}
     \\
      NAFNet~\cite{chen2022simple}  & 30.25 & 0.908 & 32.92 &  0.917   & 30.40 & 0.891  & 37.40 & 0.964 & 32.73  & 0.921
       \\
     MDARNet~\cite{MDARNet} & 28.98 & 0.892 & 33.08 & 0.919 & 29.71 & 0.884 & 35.68 & 0.961 & 31.86 &0.914
     \\
       IRNeXt~\cite{IRNeXt} &31.53 &0.919 &- & - &31.64 &0.902 & 38.14&0.972 &- &-
    \\
    SFNet~\cite{SFNet} &31.47&0.919&32.55&0.911 & \underline{31.90} & \textbf{0.908} & 38.21&0.974 &33.53&0.928
    \\
    FSNet~\cite{FSNet} &31.05&0.919 &33.08&0.916 & 31.77&\underline{0.906} &38.00 & 0.972 &33.48 &0.928
    \\
     \hline
      \textbf{MSFSNet(Ours)}  & \textbf{32.76} & \underline{0.921} & \textbf{34.99}&\textbf{0.940} & \textbf{31.98}&\textbf{0.908} &\textbf{39.01}&\textbf{0.983} &\textbf{34.69}&\textbf{0.938}
    \\
    \hline
\end{tabular}}
\end{table*}

\subsubsection{\textbf{Image Deraining}}
\begin{figure*}[htb] 
	\centering
	\includegraphics[width=1\textwidth]{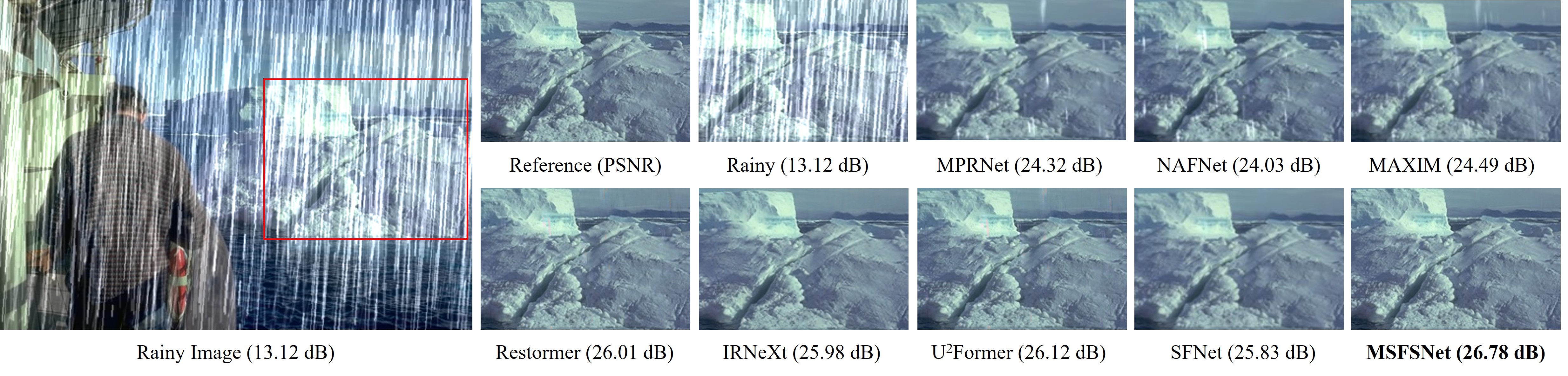}
	\caption{\textbf{Image  deraining} example on the Rain100H~\cite{Rain100}. The outputs of the MSFSNet exhibit no traces of rain streaks. }
	\label{fig:rain}
\end{figure*}

Following the prior work~\cite{Zamir2021MPRNet,MSPFN,SFNet}, we compute PSNR/SSIM scores using the Y channel in the YCbCr color space for the image deraining task. Table~\ref{tb:derain} illustrates that our method consistently outperforms existing approaches across all four datasets. Notably, our method achieves a remarkable average improvement of 0.78 dB over all datasets compared to the best-performing method Restormer~\cite{Zamir2021Restormer}. Furthermore, on the Test1200 dataset~\cite{DIDMDN}, MSFSNet exhibits a substantial 1.8 dB PSNR improvement over the previous best method Restormer. In addition to quantitative assessments, Figure~\ref{fig:rain} presents qualitative results, showcasing the effectiveness of MSFSNet in removing rain streaks of various orientations and magnitudes while preserving the structural content of the images.

\subsubsection{\textbf{ Image Denoising}}
\begin{figure*}[htb] 
	\centering
	\includegraphics[width=1\textwidth]{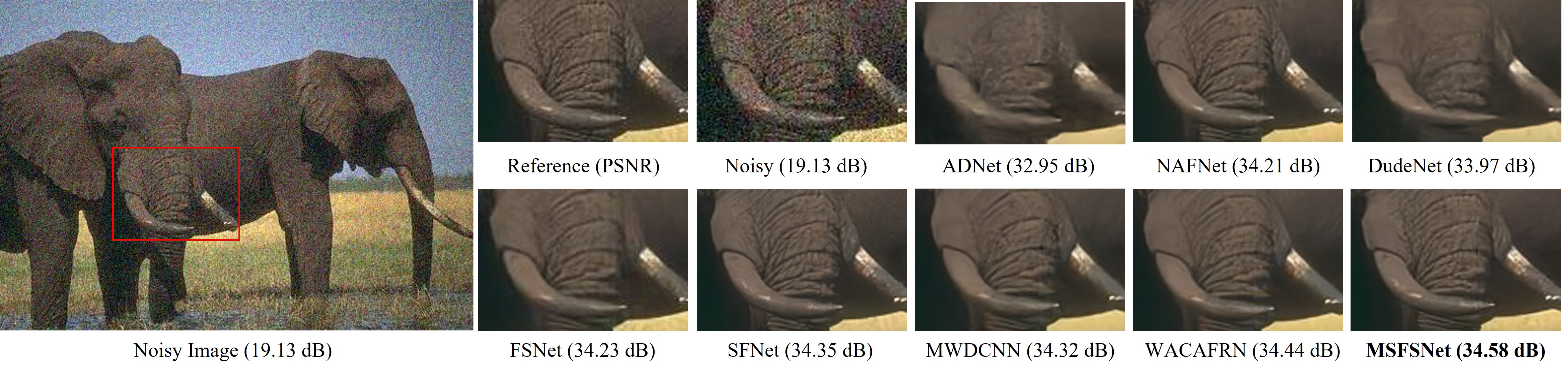}
	\caption{\textbf{Image denoising} example on the CBSD68~\cite{BSD68}. Our MSFSNet produces images that are more faithful to the ground truth in terms of quality. }
	\label{fig:noisec}
\end{figure*}

Table~\ref{tab:denoisec} showcases the performance of a single MSFSNet model across various noise levels (15, 25, and 50). Our method demonstrates considerable effectiveness across different datasets and noise levels. Notably, for the challenging noise level of 25 on the Kodak24 dataset~\cite{kodak}, MSFSNet achieves a 0.27 dB improvement over the previous best-performing method WACAFRN~\cite{WACAFRN}. Additionally, for the challenging noise level of 15 on the high-resolution Urban100 dataset~\cite{urban100}, MSFSNet exhibits a 0.19 dB improvement over SFNet~\cite{SFNet}.
In Figure~\ref{fig:noisec}, the denoised results produced by various methods are visually compared, and it is evident that our MSFSNet generates images that are more faithful to the ground truth in terms of quality.

\begin{table}
    \centering
    \caption{Image denoising results.  On Kodak24 dataset~\cite{kodak} for noise level 25, our MSFSNet is better than the state-of-the-art by 0.27 dB.}
    \label{tab:denoisec}
    \resizebox{\linewidth}{!}{
    \begin{tabular}{c|ccc|ccc|ccc}
    \hline
      & \multicolumn{3}{c|}{CBSD68~\cite{BSD68}}  & \multicolumn{3}{c|}{Kodak24~\cite{kodak}} & \multicolumn{3}{c}{Urban100~\cite{urban100}} 
    \\
   Methods  & $\sigma= $15  & $\sigma= $25 & $\sigma= $50 & $\sigma= $15  & $\sigma= $25 & $\sigma= $50 & $\sigma= $15  & $\sigma= $25 & $\sigma= $50 
   \\
   \hline
   IRCNN~\cite{IRCNN} & 33.86 &31.16 &27.86 &34.69 &32.18 &28.93 & 33.78 &31.20 &27.70
   \\
FFDNet~\cite{FFDNet} & 33.87 &31.21 &27.96 &34.63 &32.13 &28.98 & 33.83 &31.40 &28.05
\\
DnCNN~\cite{set12} & 33.90 &31.24 &27.95 &34.60 &32.14 &28.95& 32.98 &30.81 &27.59
\\
DudeNet~\cite{DudaNet} &34.01 &31.34 & 28.09 & 34.81 & 32.26 & 29.10 &- & - &-
\\
MWDCNN~\cite{MWDCNN} &34.18 & 31.45 & 28.13 & 34.92 & 32.38 & 29.27 &- &- &-
\\
WACAFRN~\cite{WACAFRN} & \underline{34.29} & \textbf{31.51} & \underline{28.19} & \underline{34.98} & \underline{32.45} & \underline{29.32}&- &- &-
\\
SFNet~\cite{SFNet} & 34.09 & \underline{31.49} & 28.02 & 34.93 & 32.42 & 29.25&\underline{34.19} &32.01 & 29.03
\\
FSNet~\cite{FSNet} & 34.11 & \textbf{31.51} & 28.01 & 34.95 & 32.42 & 29.27&34.15 & \underline{32.04} & \textbf{29.15}
\\
\hline
\textbf{MSFSNet(Ours)} & \textbf{34.42} & \textbf{31.51} & \textbf{28.26} & \textbf{35.11} & \textbf{32.72} & \textbf{29.68} &\textbf{34.38} & \textbf{32.18} & \underline{29.14}
       \\
    \hline
    \end{tabular}}
\end{table}

\subsection{Ablation Studies}
The ablation studies are conducted on image motion deblurring (GoPro~\cite{Gopro}) to analyze the impact of each of our model components. Next, we describe the impact of each component.
\begin{table}
    \caption{Ablation studies of each module.}
    \label{tab:ablem}
    \centering
    \begin{tabular}{ccccc}
    \hline
       Multi-input and Multi-output & SFF & DFS & PSNR & SSIM
       \\
       \hline
       \faTimes & \faTimes & \faTimes & 32.83 & 0.960
         \\
          \faCheck & \faTimes & \faTimes &32.92 & 0.960
          \\
           \faTimes & \faCheck & \faTimes &33.08 &0.961
           \\
           \faTimes & \faTimes & \faCheck & 33.16 & 0.963
           \\
           \faCheck & \faCheck & \faTimes &33.15 &0.962
           \\
           \faCheck & \faCheck & \faCheck &33.42 &0.964
         \\
         \hline
    \end{tabular}
\end{table}

\textbf{Effectiveness of each module.}
Table~\ref{tab:ablem} demonstrates that the baseline model achieves 32.83 dB PSNR. Multi-input and Multi-output, SFF, and DFS modules results in gains of 0.09 dB, 0.21 dB, and 0.33 dB, respectively, over the baseline model. When all modules are integrated simultaneously, our model achieves an improvement of 0.59 dB. This highlights the effectiveness of our proposed modules. 

To validate the plug-and-play nature of our SFF and DFS, we integrate them into NAFNet~\cite{chen2022simple} and Restormer~\cite{Zamir2021Restormer}.  As shown in Table~\ref{tab:ablpl}, the inclusion of SFF resulted in a 0.14 dB improvement for NAFNet and a 0.16 dB improvement for Restormer. The introduction of DFS led to performance enhancements of 0.3 dB and 0.31 dB for NAFNet and Restormer, respectively. Combining SFF and DFS together resulted in significant performance improvements of 0.44 dB and 0.42 dB for NAFNet and Restormer, respectively.

\begin{table}
    \caption{Plug-and-play ablation experiments. "w" denote with.}
    \label{tab:ablpl}
    \centering
    \resizebox{\linewidth}{!}{
    \begin{tabular}{ccc|ccc}
    \hline
    Method & PSNR & $\triangle$ PSNR  & Method & PSNR & $\triangle$ PSNR 
    \\
    \hline
    NAFNet~\cite{chen2022simple} & 32.83 & - & Restormer~\cite{Zamir2021Restormer} & 32.92 & -
    \\
     w SFF & 32.97&+0.14 &   w SFF & 33.08 &+0.16
     \\
     w DFS & 33.13 &+0.30 &w DFS & 33.23 &+0.31
     \\
     w SFF \& DFS &  33.27 & +0.44 &w SFF \& DFS & 33.35 & +0.42
     \\
         \hline
    \end{tabular}}
\end{table}

\begin{table}
    \caption{The impact of frequency selection on the overall performance.}
    \label{tab:aldfs}
    \centering
    \begin{tabular}{cccc}
    \hline
         Modules & SKFF~\cite{Zamir2022MIRNetv2} &MDSF~\cite{FSNet} & FCAM
         \\
         \hline
         PSNR   &33.21  &33.22 & 33.42
         \\ 
         \hline
         SSIM   & 0.9634 & 0.9632 &0.9643
         \\
         \hline
    \end{tabular}
\end{table}

\begin{figure*}[htb] 
	\centering
	\includegraphics[width=1\linewidth]{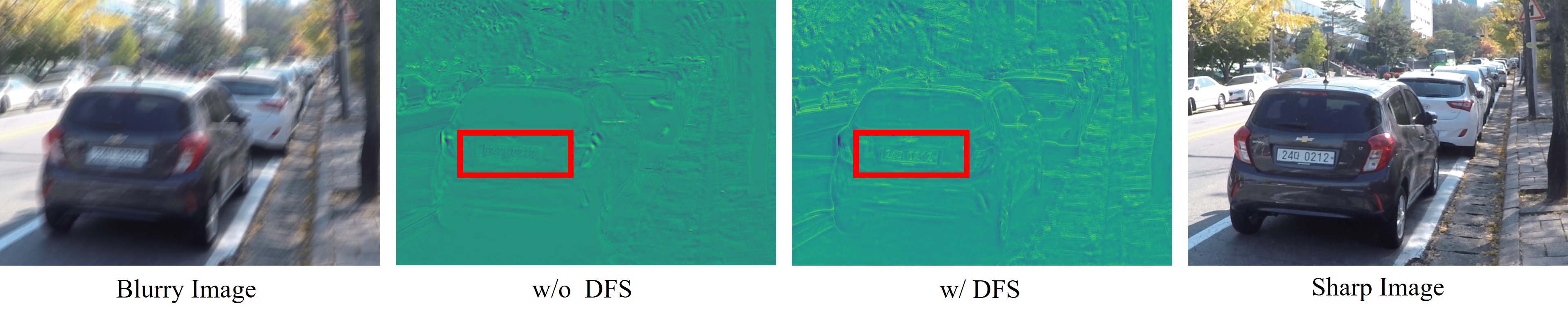}
	\caption{Visualization of intermediate feature maps of models
with and without DFS.}
	\label{fig:wdfs}
\end{figure*}

\begin{figure}[htb] 
	\centering
	\includegraphics[width=1\linewidth]{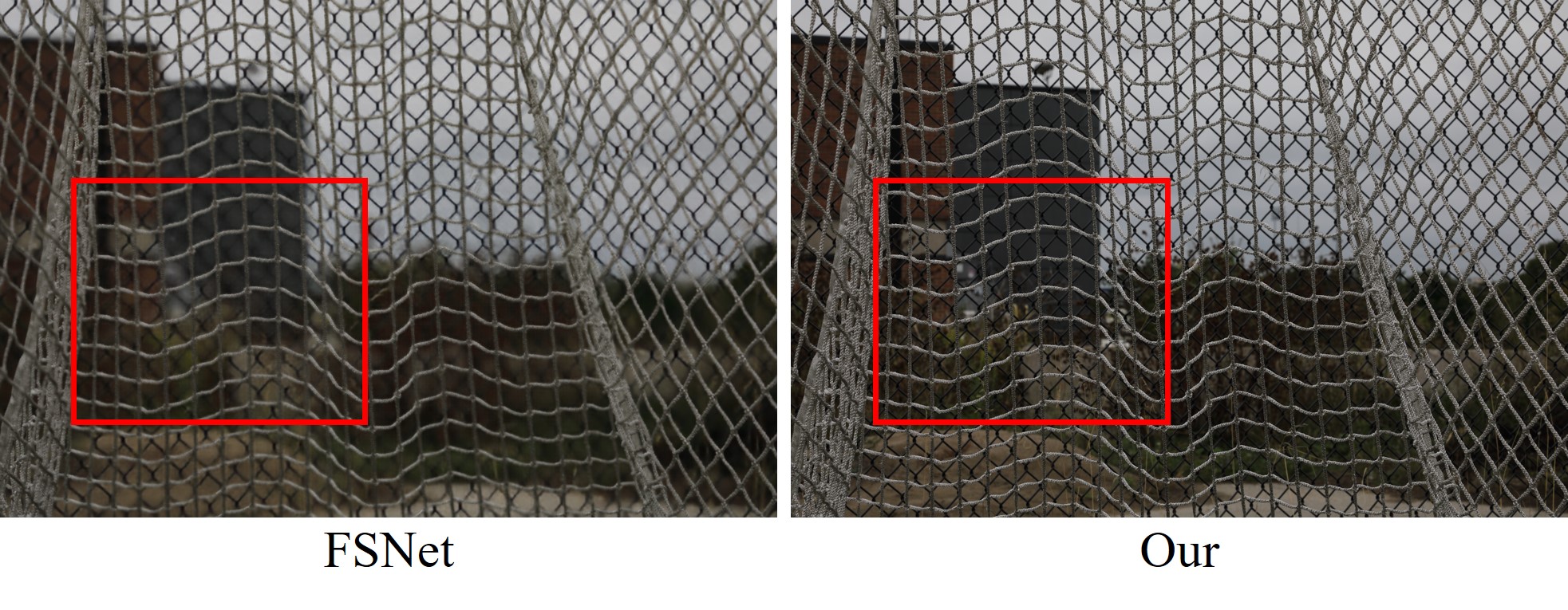}
	\caption{Visualization compare for defocus deblurring.}
	\label{fig:bgs}
\end{figure}

\begin{table}
    \caption{The impact of group number on the overall performance.}
    \label{tab:aldfsg}
    \centering
    \begin{tabular}{ccccc}
    \hline
         Group&  2&  4  & 8 & 16
         \\
         \hline
         PSNR&  33.35  &33.38  &33.42&33.31
         \\ 
         \hline
    \end{tabular}
\end{table}

\textbf{Design choices for DFS.} To examine the advantage of our DFS design, we compare our RCAM within DFS with other frequency selection method, such as SKFF~\cite{Zamir2022MIRNetv2} and MDSF~\cite{FSNet}. As shown in Table~\ref{tab:aldfs},  our RCAM achieves better results, which indicates that our RCAM is able to select the most useful frequencies for fusion, enabling our model to select the most informative components for recovery. In particular, our approach exhibits a 0.2 dB improvement in performance when compared to MDSF~\cite{FSNet}. This suggests that our method is more effective in accurately emphasizing the relevant components for reconstruction as compared to the modulator employed by MDSF. 
In addition, we also conducted experiments on group selection, as shown in Table~\ref{tab:aldfsg}, and achieved better results when 8 was selected. To delve into the DFS, we  plot resulting feature maps of MSFSBlock for models with and without DFS in Figure~\ref{fig:wdfs}.  As can be seen, equipped with DFS, the model recovers  more details of the number plate than that without DFS. We also show the visual comparison with FSNet~\cite{FSNet}, as shown by the red box in Figure~\ref{fig:bgs}, FSNet uses binary frequency decomposition
and cannot strike a balance between overlapped scenes, while our method  can solve this problem and provide more detailed information.

\begin{table}
    \caption{The impact of skip connections on the overall performance.}
    \label{tab:rwal}
    \centering
    \begin{tabular}{ccccc}
    \hline
         Modules&  Sum&  Concatenation & CGB~\cite{tu2022maxim} & SFF
         \\
         \hline
         PSNR&  33.23  &33.25  &33.27&33.42
         \\ 
         \hline
         SSIM   & 0.9628& 0.9637 & 0.9641 &0.9643
         \\
         \hline
    \end{tabular}
\end{table}

\begin{table}
    \centering
    \caption{The evaluation of model computational complexity on the GoPro dataset~\cite{Gopro}.}
    \label{tab:computational}
    \begin{tabular}{ccccc}
    \hline
         Method& Time(s) & MACs(G)  & PSNR & SSIM
         \\
         \hline\hline
         DBGAN~\cite{DBGAN}& 1.447 & 759 & 31.10 &0.942
         \\
         MPRNet~\cite{Zamir2021MPRNet} & 1.148 & 777 & 32.66 &0.959
         \\
         Restormer~\cite{Zamir2021Restormer} & 1.218 & 140 & 32.92 & 0.961
         \\
         Stripformer~\cite{Tsai2022Stripformer} &1.054 &170 & 33.08 &0.962
         \\
         IRNeXt~\cite{IRNeXt} &\underline{0.255} & 114 & 33.16 & 0.962
         \\
         SFNet~\cite{SFNet} &0.408 & 125 & 33.27 &\underline{0.963}
         \\
         FSNet~\cite{FSNet} &0.362 & \underline{111}& \underline{33.29} & \underline{0.963}
         \\
         \hline
         MSFSNet(Ours) &\textbf{0.244} &\textbf{27} & \textbf{33.42} & \textbf{0.964}
         \\
         \hline
    \end{tabular}

\end{table}
\textbf{Design choices for skip connections.} As shown in Table~\ref{tab:rwal}, we compare our SFF with several skip connections alternatives. 
In the initial stage, we utilize conventional addition and concatenation techniques for skip connections, yielding PSNR values of 33.23 dB and 33.25 dB, respectively. Subsequently, we integrate CGB~\cite{tu2022maxim} to regulate information flow within the skip connection, resulting in a PSNR improvement to 33.27 dB. When compared to these approaches, our proposed SFF surpasses them, achieving even higher results at 33.42 dB.  
This suggests that our SFF effectively merges features across various scales by discerning the information transmitted through skip connections. It retains their distinct complementary properties, allowing the model to acquire a comprehensive and precise feature set.

\subsection{Resource Efficient}
As deep learning models pursue higher accuracy, their complexity rises, necessitating more resources. However, in certain contexts, deploying larger models may be impractical. Table~\ref{tab:computational} and Figure~\ref{fig:param} demonstrate that our MSFSNet model attains state-of-the-art performance while concurrently reducing computational costs. Specifically, on the GoPro dataset~\cite{Gopro}, we achieve a 0.13 dB improvement over the previous best approach, FSNet~\cite{FSNet}, while using only a quarter of the computational resources and achieving nearly 1.5$\times$ faster inference. This highlights the efficiency of our method, showcasing superior performance along with resource effectiveness.

\section{Conclusion}
In this paper, we present a multi-scale frequency selection network (MSFSNet) comprising two seamlessly integrable plug-in modules designed for incorporation into any existing restoration network. Specifically, a dynamic filtering selection module (DFS) is devised to generate high and low-frequency information, discern which information to retain through the frequency cross-attention mechanism (FCAM), and select the most informative component for recovery. Additionally, to glean a more multi-scale and accurate set of hybrid features, we introduce a skip feature fusion block (SFF) that discriminatively determines the information to be propagated in skip connections. Extensive experiments showcase that our MSFSNet achieves state-of-the-art performance across four typical image restoration tasks.

\bibliographystyle{IEEEtran}
\bibliography{refbib}

\begin{thebibliography}{10}
\providecommand{\url}[1]{#1}
\csname url@samestyle\endcsname
\providecommand{\newblock}{\relax}
\providecommand{\bibinfo}[2]{#2}
\providecommand{\BIBentrySTDinterwordspacing}{\spaceskip=0pt\relax}
\providecommand{\BIBentryALTinterwordstretchfactor}{4}
\providecommand{\BIBentryALTinterwordspacing}{\spaceskip=\fontdimen2\font plus
\BIBentryALTinterwordstretchfactor\fontdimen3\font minus \fontdimen4\font\relax}
\providecommand{\BIBforeignlanguage}[2]{{%
\expandafter\ifx\csname l@#1\endcsname\relax
\typeout{** WARNING: IEEEtran.bst: No hyphenation pattern has been}%
\typeout{** loaded for the language `#1'. Using the pattern for}%
\typeout{** the default language instead.}%
\else
\language=\csname l@#1\endcsname
\fi
#2}}
\providecommand{\BIBdecl}{\relax}
\BIBdecl

\bibitem{yang2020single}
W.~Yang, R.~T. Tan, S.~Wang, Y.~Fang, and J.~Liu, ``Single image deraining: From model-based to data-driven and beyond,'' \emph{IEEE Transactions on pattern analysis and machine intelligence}, vol.~43, no.~11, pp. 4059--4077, 2020.

\bibitem{karaali2017edge}
A.~Karaali and C.~R. Jung, ``Edge-based defocus blur estimation with adaptive scale selection,'' \emph{IEEE Transactions on Image Processing}, vol.~27, no.~3, pp. 1126--1137, 2017.

\bibitem{2011Image}
W.~Dong, L.~Zhang, G.~Shi, and X.~Wu, ``Image deblurring and super-resolution by adaptive sparse domain selection and adaptive regularization,'' \emph{IEEE Transactions on Image Processing}, vol.~20, no.~7, pp. 1838--1857, 2011.

\bibitem{2011Single}
K.~He, J.~Sun, and X.~Tang, ``Single image haze removal using dark channel prior.'' \emph{IEEE Transactions on Pattern Analysis and Machine Intelligence}, 2011.

\bibitem{IRNeXt}
Y.~Cui, W.~Ren, S.~Yang, X.~Cao, and A.~Knoll, ``Irnext: Rethinking convolutional network design for image restoration,'' in \emph{Proceedings of the 40th International Conference on Machine Learning}, 2023.

\bibitem{chen2022simple}
L.~Chen, X.~Chu, X.~Zhang, and J.~Sun, ``Simple baselines for image restoration,'' \emph{ECCV}, 2022.

\bibitem{2021Rethinking}
S.~J. Cho, S.~W. Ji, J.~P. Hong, S.~W. Jung, and S.~J. Ko, ``Rethinking coarse-to-fine approach in single image deblurring,'' in \emph{ICCV}, 2021.

\bibitem{Zamir2021MPRNet}
S.~W. Zamir, A.~Arora, S.~Khan, M.~Hayat, F.~S. Khan, M.-H. Yang, and L.~Shao, ``Multi-stage progressive image restoration,'' in \emph{CVPR}, 2021.

\bibitem{Chen_2021_CVPR}
L.~Chen, X.~Lu, J.~Zhang, X.~Chu, and C.~Chen, ``Hinet: Half instance normalization network for image restoration,'' in \emph{Proceedings of the IEEE/CVF Conference on Computer Vision and Pattern Recognition (CVPR) Workshops}, June 2021, pp. 182--192.

\bibitem{PREnet}
D.~Ren, W.~Zuo, Q.~Hu, P.~F. Zhu, and D.~Meng, ``Progressive image deraining networks: A better and simpler baseline,'' \emph{2019 IEEE/CVF Conference on Computer Vision and Pattern Recognition (CVPR)}, pp. 3932--3941, 2019.

\bibitem{RESCAN}
X.~Li, J.~Wu, Z.~Lin, H.~Liu, and H.~Zha, ``Recurrent squeeze-and-excitation context aggregation net for single image deraining,'' in \emph{European Conference on Computer Vision}, 2018.

\bibitem{2018LearningD}
J.~Pan, S.~Liu, D.~Sun, J.~Zhang, Y.~Liu, J.~Ren, Z.~Li, J.~Tang, H.~Lu, and Y.~W.~a. Tai, ``Learning dual convolutional neural networks for low-level vision,'' in \emph{CVPR}, 2018.

\bibitem{2022Learning}
J.~Pan, D.~Sun, J.~Zhang, J.~Tang, J.~Yang, Y.~W. Tai, and M.~H. Yang, ``Dual convolutional neural networks for low-level vision,'' \emph{International Journal of Computer Vision}, 2022.

\bibitem{2020Refining}
V.~Singh, K.~Ramnath, and A.~Mittal, ``Refining high-frequencies for sharper super-resolution and deblurring,'' \emph{Computer Vision and Image Understanding}, vol. 199, no.~C, p. 103034, 2020.

\bibitem{chen2020decomposition}
D.~Chen and M.~E. Davies, ``Deep decomposition learning for inverse imaging problems,'' in \emph{Proceedings of the European Conference on Computer Vision (ECCV)}, 2020.

\bibitem{zhang2018image}
Y.~Zhang, K.~Li, K.~Li, L.~Wang, B.~Zhong, and Y.~Fu, ``Image super-resolution using very deep residual channel attention networks,'' in \emph{Proceedings of the European Conference on Computer Vision (ECCV)}, 2018, pp. 286--301.

\bibitem{2019Real}
S.~Anwar, ``Real image denoising with feature attention,'' \emph{ICCV}, 2019.

\bibitem{ms9919385}
L.~Zheng, Y.~Li, K.~Zhang, and W.~Luo, ``T-net: Deep stacked scale-iteration network for image dehazing,'' \emph{IEEE Transactions on Multimedia}, vol.~25, pp. 6794--6807, 2023.

\bibitem{Degan}
O.~Kupyn, V.~Budzan, M.~Mykhailych, D.~Mishkin, and J.~Matas, ``Deblurgan: Blind motion deblurring using conditional adversarial networks,'' \emph{2018 IEEE/CVF Conference on Computer Vision and Pattern Recognition}, pp. 8183--8192, 2017.

\bibitem{DBGAN}
K.~Zhang, W.~Luo, Y.~Zhong, L.~Ma, B.~Stenger, W.~Liu, and H.~Li, ``Deblurring by realistic blurring,'' \emph{2020 IEEE/CVF Conference on Computer Vision and Pattern Recognition (CVPR)}, pp. 2734--2743, 2020.

\bibitem{deganv2}
O.~Kupyn, T.~Martyniuk, J.~Wu, and Z.~Wang, ``Deblurgan-v2: Deblurring (orders-of-magnitude) faster and better,'' \emph{2019 IEEE/CVF International Conference on Computer Vision (ICCV)}, pp. 8877--8886, 2019.

\bibitem{mg10130403}
K.~Wu, J.~Huang, Y.~Ma, F.~Fan, and J.~Ma, ``Cycle-retinex: Unpaired low-light image enhancement via retinex-inline cyclegan,'' \emph{IEEE Transactions on Multimedia}, pp. 1--16, 2023.

\bibitem{mg9671019}
S.~Wu, C.~Dong, and Y.~Qiao, ``Blind image restoration based on cycle-consistent network,'' \emph{IEEE Transactions on Multimedia}, vol.~25, pp. 1111--1124, 2023.

\bibitem{u2former}
X.~Feng, H.~Ji, W.~Pei, J.~Li, G.~Lu, and D.~Zhang, ``U2-former: Nested u-shaped transformer for image restoration via multi-view contrastive learning,'' \emph{IEEE Transactions on Circuits and Systems for Video Technology}, pp. 1--1, 2023.

\bibitem{IDT}
J.~Xiao, X.~Fu, A.~Liu, F.~Wu, and Z.-J. Zha, ``Image de-raining transformer,'' \emph{IEEE Transactions on Pattern Analysis and Machine Intelligence}, vol.~45, no.~11, pp. 12\,978--12\,995, 2023.

\bibitem{Zamir2021Restormer}
S.~W. Zamir, A.~Arora, S.~Khan, M.~Hayat, F.~S. Khan, and M.-H. Yang, ``Restormer: Efficient transformer for high-resolution image restoration,'' in \emph{CVPR}, 2022.

\bibitem{Tsai2022Stripformer}
F.-J. Tsai, Y.-T. Peng, Y.-Y. Lin, C.-C. Tsai, and C.-W. Lin, ``Stripformer: Strip transformer for fast image deblurring,'' in \emph{ECCV}, 2022.

\bibitem{Wang_2022_CVPR}
Z.~Wang, X.~Cun, J.~Bao, W.~Zhou, J.~Liu, and H.~Li, ``Uformer: A general u-shaped transformer for image restoration,'' in \emph{Proceedings of the IEEE/CVF Conference on Computer Vision and Pattern Recognition (CVPR)}, June 2022, pp. 17\,683--17\,693.

\bibitem{mt10387581}
X.~Zhou, H.~Huang, Z.~Wang, and R.~He, ``Ristra: Recursive image super-resolution transformer with relativistic assessment,'' \emph{IEEE Transactions on Multimedia}, pp. 1--12, 2024.

\bibitem{kong2023efficient}
L.~Kong, J.~Dong, J.~Ge, M.~Li, and J.~Pan, ``Efficient frequency domain-based transformers for high-quality image deblurring,'' in \emph{Proceedings of the IEEE/CVF Conference on Computer Vision and Pattern Recognition}, 2023, pp. 5886--5895.

\bibitem{fLi2023ICLR}
C.~Li, C.-L. Guo, M.~Zhou, Z.~Liang, S.~Zhou, R.~Feng, and C.~C. Loy, ``Embedding fourier for ultra-high-definition low-light image enhancement,'' in \emph{ICLR}, 2023.

\bibitem{f8803391}
H.-H. Yang and Y.~Fu, ``Wavelet u-net and the chromatic adaptation transform for single image dehazing,'' in \emph{2019 IEEE International Conference on Image Processing (ICIP)}, 2019, pp. 2736--2740.

\bibitem{fxint2023freqsel}
M.~Xintian, L.~Yiming, L.~Fengze, L.~Qingli, S.~Wei, and W.~Yan, ``Intriguing findings of frequency selection for image deblurring,'' in \emph{Proceedings of the 37th AAAI Conference on Artificial Intelligence}, 2023.

\bibitem{WACAFRN}
S.~Ding, Q.~Wang, L.~Guo, X.~Li, L.~Ding, and X.~Wu, ``Wavelet and adaptive coordinate attention guided fine-grained residual network for image denoising,'' \emph{IEEE Transactions on Circuits and Systems for Video Technology}, 2024.

\bibitem{huang2022WINNet}
J.-J. Huang and P.~L. Dragotti, ``{WINNet}: Wavelet-inspired invertible network for image denoising,'' \emph{IEEE Transactions on Image Processing}, vol.~31, pp. 4377--4392, 2022.

\bibitem{mf9786841}
W.~Zou, L.~Chen, Y.~Wu, Y.~Zhang, Y.~Xu, and J.~Shao, ``Joint wavelet sub-bands guided network for single image super-resolution,'' \emph{IEEE Transactions on Multimedia}, vol.~25, pp. 4623--4637, 2023.

\bibitem{mf9917526}
Z.~Sheng, X.~Liu, S.-Y. Cao, H.-L. Shen, and H.~Zhang, ``Frequency-domain deep guided image denoising,'' \emph{IEEE Transactions on Multimedia}, vol.~25, pp. 6767--6781, 2023.

\bibitem{FSNet}
Y.~Cui, W.~Ren, X.~Cao, and A.~Knoll, ``Image restoration via frequency selection,'' \emph{IEEE Transactions on Pattern Analysis and Machine Intelligence}, vol.~46, no.~2, pp. 1093--1108, 2024.

\bibitem{SFNet}
Y.~Cui, Y.~Tao, Z.~Bing, W.~Ren, X.~Gao, X.~Cao, K.~Huang, and A.~Knoll, ``Selective frequency network for image restoration,'' in \emph{The Eleventh International Conference on Learning Representations}, 2023.

\bibitem{10196308}
T.~Gao, Y.~Wen, K.~Zhang, J.~Zhang, T.~Chen, L.~Liu, and W.~Luo, ``Frequency-oriented efficient transformer for all-in-one weather-degraded image restoration,'' \emph{IEEE Transactions on Circuits and Systems for Video Technology}, pp. 1--1, 2023.

\bibitem{2013Unnatural}
X.~Li, S.~Zheng, and J.~Jia, ``Unnatural l0 sparse representation for natural image deblurring,'' in \emph{IEEE Conference on Computer Vision and Pattern Recognition}, 2013.

\bibitem{ffanet}
X.~Qin, Z.~Wang, Y.~Bai, X.~Xie, and H.~Jia, ``Ffa-net: Feature fusion attention network for single image dehazing,'' in \emph{Proceedings of the AAAI Conference on Artificial Intelligence}, vol.~34, no.~07, 2020, pp. 11\,908--11\,915.

\bibitem{Zamir2022MIRNetv2}
S.~W. Zamir, A.~Arora, S.~Khan, M.~Hayat, F.~S. Khan, M.-H. Yang, and L.~Shao, ``Learning enriched features for fast image restoration and enhancement,'' \emph{IEEE Transactions on Pattern Analysis and Machine Intelligence (TPAMI)}, 2022.

\bibitem{2017Attention}
A.~Vaswani, N.~Shazeer, N.~Parmar, J.~Uszkoreit, L.~Jones, A.~N. Gomez, L.~Kaiser, and I.~Polosukhin, ``Attention is all you need,'' \emph{arXiv}, 2017.

\bibitem{IPT}
H.~Chen, Y.~Wang, T.~Guo, C.~Xu, Y.~Deng, Z.~Liu, S.~Ma, C.~Xu, C.~Xu, and W.~Gao, ``Pre-trained image processing transformer,'' \emph{2021 IEEE/CVF Conference on Computer Vision and Pattern Recognition (CVPR)}, pp. 12\,294--12\,305, 2020.

\bibitem{liang2021swinir}
J.~Liang, J.~Cao, G.~Sun, K.~Zhang, L.~Van~Gool, and R.~Timofte, ``Swinir: Image restoration using swin transformer,'' \emph{arXiv preprint arXiv:2108.10257}, 2021.

\bibitem{Chi_2020_FFC}
L.~Chi, B.~Jiang, and Y.~Mu, ``Fast fourier convolution,'' in \emph{Advances in Neural Information Processing Systems}, 2020.

\bibitem{Gopro}
S.~Nah, T.~H. Kim, and K.~M. Lee, ``Deep multi-scale convolutional neural network for dynamic scene deblurring,'' \emph{2017 IEEE Conference on Computer Vision and Pattern Recognition (CVPR)}, pp. 257--265, 2016.

\bibitem{HIDE}
Z.~Shen, W.~Wang, X.~Lu, J.~Shen, H.~Ling, T.~Xu, and L.~Shao, ``Human-aware motion deblurring,'' \emph{2019 IEEE/CVF International Conference on Computer Vision (ICCV)}, pp. 5571--5580, 2019.

\bibitem{realblurrim_2020_ECCV}
J.~Rim, H.~Lee, J.~Won, and S.~Cho, ``Real-world blur dataset for learning and benchmarking deblurring algorithms,'' in \emph{Proceedings of the European Conference on Computer Vision (ECCV)}, 2020.

\bibitem{DPDNet}
A.~Abuolaim and M.~S. Brown, ``Defocus deblurring using dual-pixel data,'' in \emph{European Conference on Computer Vision}.\hskip 1em plus 0.5em minus 0.4em\relax Springer, 2020, pp. 111--126.

\bibitem{Rain100}
W.~Yang, R.~T. Tan, J.~Feng, J.~Liu, Z.~Guo, and S.~Yan, ``Deep joint rain detection and removal from a single image,'' \emph{2017 IEEE Conference on Computer Vision and Pattern Recognition (CVPR)}, pp. 1685--1694, 2016.

\bibitem{Test100}
H.~Zhang, V.~A. Sindagi, and V.~M. Patel, ``Image de-raining using a conditional generative adversarial network,'' \emph{IEEE Transactions on Circuits and Systems for Video Technology}, vol.~30, pp. 3943--3956, 2017.

\bibitem{8099669}
X.~Fu, J.~Huang, D.~Zeng, Y.~Huang, X.~Ding, and J.~Paisley, ``Removing rain from single images via a deep detail network,'' in \emph{2017 IEEE Conference on Computer Vision and Pattern Recognition (CVPR)}, 2017, pp. 1715--1723.

\bibitem{7780668}
Y.~Li, R.~T. Tan, X.~Guo, J.~Lu, and M.~S. Brown, ``Rain streak removal using layer priors,'' in \emph{2016 IEEE Conference on Computer Vision and Pattern Recognition (CVPR)}, 2016, pp. 2736--2744.

\bibitem{DIDMDN}
H.~Zhang and V.~M. Patel, ``Density-aware single image de-raining using a multi-stream dense network,'' \emph{2018 IEEE/CVF Conference on Computer Vision and Pattern Recognition}, pp. 695--704, 2018.

\bibitem{DIK}
E.~Agustsson and R.~Timofte, ``Ntire 2017 challenge on single image super-resolution: Dataset and study,'' in \emph{2017 IEEE Conference on Computer Vision and Pattern Recognition Workshops (CVPRW)}, 2017, pp. 1122--1131.

\bibitem{lim2017enhanced}
B.~Lim, S.~Son, H.~Kim, S.~Nah, and K.~Mu~Lee, ``Enhanced deep residual networks for single image super-resolution,'' in \emph{Proceedings of the IEEE conference on computer vision and pattern recognition workshops}, 2017, pp. 136--144.

\bibitem{BSD500}
P.~Arbeláez, M.~Maire, C.~Fowlkes, and J.~Malik, ``Contour detection and hierarchical image segmentation,'' \emph{IEEE Transactions on Pattern Analysis and Machine Intelligence}, vol.~33, no.~5, pp. 898--916, 2011.

\bibitem{ma2016waterloo}
K.~Ma, Z.~Duanmu, Q.~Wu, Z.~Wang, H.~Yong, H.~Li, and L.~Zhang, ``Waterloo exploration database: New challenges for image quality assessment models,'' \emph{IEEE Transactions on Image Processing}, vol.~26, no.~2, pp. 1004--1016, 2016.

\bibitem{BSD68}
D.~Martin, C.~Fowlkes, D.~Tal, and J.~Malik, ``A database of human segmented natural images and its application to evaluating segmentation algorithms and measuring ecological statistics,'' in \emph{Proceedings Eighth IEEE International Conference on Computer Vision. ICCV 2001}, vol.~2, 2001, pp. 416--423 vol.2.

\bibitem{urban100}
J.-B. Huang, A.~Singh, and N.~Ahuja, ``Single image super-resolution from transformed self-exemplars,'' in \emph{2015 IEEE Conference on Computer Vision and Pattern Recognition (CVPR)}, 2015, pp. 5197--5206.

\bibitem{kodak}
R.~Franzen, ``Kodak lossless true color image suite,'' \emph{source: http://r0k. us/graphics/kodak}, vol.~4, no.~2, p.~9, 1999.

\bibitem{SPAIR}
K.~Purohit, M.~Suin, A.~N. Rajagopalan, and V.~N. Boddeti, ``Spatially-adaptive image restoration using distortion-guided networks,'' \emph{CoRR}, vol. abs/2108.08617, 2021.

\bibitem{MSFSnet}
Y.~Zhang, Q.~Li, M.~Qi, D.~Liu, J.~Kong, and J.~Wang, ``Multi-scale frequency separation network for image deblurring,'' \emph{IEEE Transactions on Circuits and Systems for Video Technology}, vol.~33, no.~10, pp. 5525--5537, 2023.

\bibitem{JNB}
J.~Shi, L.~Xu, and J.~Jia, ``Just noticeable defocus blur detection and estimation,'' in \emph{2015 IEEE Conference on Computer Vision and Pattern Recognition (CVPR)}, 2015, pp. 657--665.

\bibitem{KPAC}
H.~Son, J.~Lee, S.~Cho, and S.~Lee, ``Single image defocus deblurring using kernel-sharing parallel atrous convolutions,'' in \emph{Proceedings of the IEEE/CVF International Conference on Computer Vision}, 2021, pp. 2642--2650.

\bibitem{IFAN}
J.~Lee, H.~Son, J.~Rim, S.~Cho, and S.~Lee, ``Iterative filter adaptive network for single image defocus deblurring,'' in \emph{Proceedings of the IEEE/CVF Conference on Computer Vision and Pattern Recognition}, 2021, pp. 2034--2042.

\bibitem{2014Adam}
D.~Kingma and J.~Ba, ``Adam: A method for stochastic optimization,'' \emph{Computer Science}, 2014.

\bibitem{2016SGDR}
I.~Loshchilov and F.~Hutter, ``Sgdr: Stochastic gradient descent with warm restarts,'' 2016.

\bibitem{MDARNet}
Z.~Hao, S.~Gai, and P.~Li, ``Multi-scale self-calibrated dual-attention lightweight residual dense deraining network based on monogenic wavelets,'' \emph{IEEE Transactions on Circuits and Systems for Video Technology}, vol.~33, no.~6, pp. 2642--2655, 2023.

\bibitem{MSPFN}
K.~Jiang, Z.~Wang, P.~Yi, C.~Chen, B.~Huang, Y.~Luo, J.~Ma, and J.~Jiang, ``Multi-scale progressive fusion network for single image deraining,'' \emph{CVPR}, 2020.

\bibitem{IRCNN}
K.~Zhang, W.~Zuo, S.~Gu, and L.~Zhang, ``Learning deep cnn denoiser prior for image restoration,'' in \emph{2017 IEEE Conference on Computer Vision and Pattern Recognition (CVPR)}, 2017, pp. 2808--2817.

\bibitem{FFDNet}
K.~Zhang, W.~Zuo, and L.~Zhang, ``Ffdnet: Toward a fast and flexible solution for {CNN} based image denoising,'' \emph{IEEE Transactions on Image Processing}, 2018.

\bibitem{set12}
K.~Zhang, W.~Zuo, Y.~Chen, D.~Meng, and L.~Zhang, ``Beyond a gaussian denoiser: Residual learning of deep cnn for image denoising,'' \emph{IEEE Transactions on Image Processing}, vol.~26, no.~7, pp. 3142--3155, 2017.

\bibitem{DudaNet}
C.~Tian, Y.~Xu, W.~Zuo, B.~Du, C.-W. Lin, and D.~Zhang, ``Designing and training of a dual cnn for image denoising,'' \emph{Knowledge-Based Systems}, vol. 226, p. 106949, 2021.

\bibitem{MWDCNN}
C.~Tian, M.~Zheng, W.~Zuo, B.~Zhang, Y.~Zhang, and D.~Zhang, ``Multi-stage image denoising with the wavelet transform,'' \emph{Pattern Recognition}, vol. 134, p. 109050, 2023.

\bibitem{tu2022maxim}
Z.~Tu, H.~Talebi, H.~Zhang, F.~Yang, P.~Milanfar, A.~Bovik, and Y.~Li, ``Maxim: Multi-axis mlp for image processing,'' \emph{CVPR}, 2022.

\end{thebibliography}

\vfill

\end{document}